%% file: main.tex
\documentclass{article}
\usepackage{iclr2027_conference}
\input{math_commands.tex}

\usepackage{booktabs}
\usepackage{array}
\usepackage{graphicx}
\usepackage{placeins}
\usepackage{xcolor}
\usepackage{hyperref}
\usepackage{float}
\usepackage{url}
\usepackage{makecell}
\definecolor{linkc}{rgb}{0, 0.44, 0.74}
\definecolor{eqc}{rgb}{1, 0, 0}
\hypersetup{
  breaklinks=true,
  colorlinks=true,
  linkcolor=eqc,
  citecolor=linkc,
  urlcolor=eqc,
  pdftitle={Honeycomb: Constant-Size Scene Memory Representation for Video World Models},
  pdfauthor={Jack Wei Lun Shi, Kaichen Zhou, Haoyu Chen, Yufeng Weng, Keane Ong, Ruojin Cai, Hang Hua, Justin K.W. Yeoh, Mengyu Wang},
  pdfsubject={Video world models},
  pdfkeywords={video world models, memory, latent spatial memory, HexMemory, HexPlane, long-horizon video generation}
}

\title{\centering{Honeycomb: Constant-Size Scene Memory \\ Representation for Video World Models\vspace{1em}}}

\author{%
\hspace*{-\tabcolsep}\makebox[\textwidth][c]{%
\begin{tabular}{c}
\textbf{Jack Wei Lun Shi$^{2,\ast}$\quad%
Kaichen Zhou$^{1,3,\ast,\dagger}$} \\[0.3em]
\textbf{Haoyu Chen$^{1}$\quad%
Yufeng Weng$^{2}$\quad%
Keane Ong$^{2,3}$\quad%
Ruojin Cai$^{1}$\quad%
Hang Hua$^{4}$} \\[0.3em]
\textbf{Justin K.W. Yeoh$^{2}$\quad%
Mengyu Wang$^{1}$} \\[0.6em]
\normalfont
$^{\ast}$Equal Contribution \quad
$^{\dagger}$Project Lead \\[0.4em]
\normalfont
$^{1}$Harvard University \quad
$^{2}$National University of Singapore \\[0.3em]
\normalfont
$^{3}$MIT \quad
$^{4}$MIT-IBM Watson AI Lab
\end{tabular}}%
}

\iclrfinalcopy
\newcommand{\method}{Honeycomb}
\newcommand{\hexmem}{\mathcal{H}}

\newcommand{\RR}{\mathbb{R}}
\newcolumntype{L}[1]{>{\raggedright\arraybackslash}p{#1}}

\renewcommand{\headrulewidth}{0pt}

\begin{document}
\maketitle
\begingroup
\renewcommand{\thefootnote}{}
\footnotetext{
  \parbox[t]{\linewidth}{
    % \textsuperscript{*}Equal contribution as first authors.\\
    Code repo: \href{https://github.com/kaichen-z/honeycomb}{Github}
    
  }
}
\endgroup

\pagestyle{plain}
\thispagestyle{plain} 

\begin{figure}[h!]
\vspace{-1em}
\centering
\includegraphics[width=\textwidth]{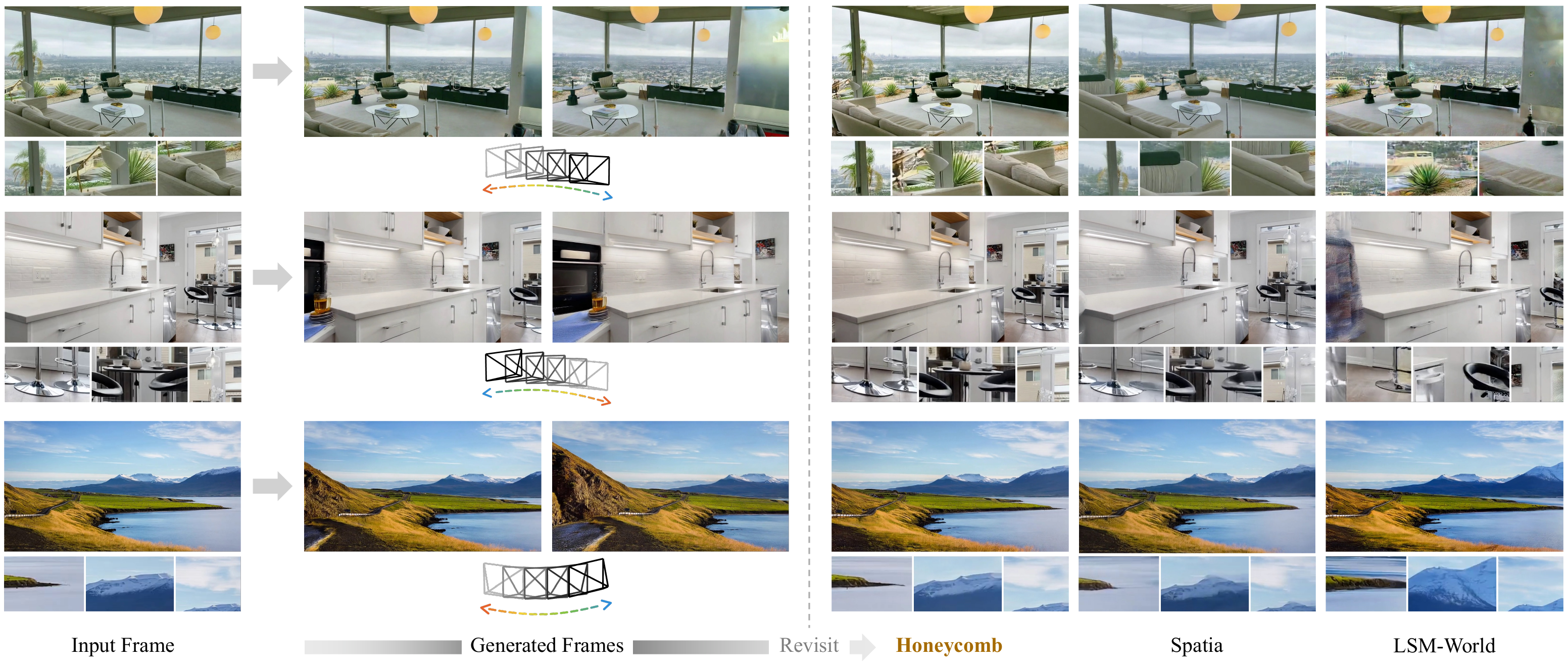}
\vspace{-1em}
\caption{\textbf{Revisiting a scene with constant-size memory.}
Given a single input frame and a camera trajectory that moves away and returns to the initial pose, \method\ generates a video whose final frame is consistent with the input frame.}
\label{fig:teaser}
\end{figure}

\begin{abstract}
Video world models require persistent scene memory to maintain consistency during long-horizon video generation. Existing spatial memories accumulate RGB observations or latent features, increasing storage requirements as generation proceeds. We introduce \method,  a video world model built on HexMemory, our proposed low-rank representation for storing scene features in a fixed-size memory with a total of six spatial and spatiotemporal planes. A feed-forward writer maps each generated chunk into new plane features. As the spatial coverage or temporal range expands, we warp the previous planes while preserving their dimensions, then fuse them with the new features through confidence-weighted pooling and a learned residual correction. A reader retrieves latents from HexMemory to condition subsequent video generation. The writer processes only observations from the new chunk, avoiding per-scene optimization and repeated processing of the full history. Experiments on WorldScore and RealEstate10K demonstrate strong video generation quality and robust revisit consistency while keeping HexMemory feature storage constant throughout generation. Code and additional visualizations are available on our \href{https://jackswl.github.io/honeycomb/}{project page}.
\end{abstract}

\input{section/1_intro}
\input{section/2_related}
\input{section/3_method}
\input{section/4_exp}
\input{section/5_conclusion}
% \clearpage
\bibliography{iclr2027_conference}
\bibliographystyle{iclr2027_conference}

\clearpage
\input{section/6_appendix}
\input{section/7_qualitative_figures}

\end{document}

%% file: math_commands.tex
\usepackage{amsmath,amsfonts,bm}

\def\eqref#1{equation~\ref{#1}}
\def\1{\bm{1}}

\def\vc{{\bm{c}}}
\def\vd{{\bm{d}}}

\def\vo{{\bm{o}}}
\def\vp{{\bm{p}}}

\def\vz{{\bm{z}}}

\DeclareMathAlphabet{\mathsfit}{\encodingdefault}{\sfdefault}{m}{sl}
\SetMathAlphabet{\mathsfit}{bold}{\encodingdefault}{\sfdefault}{bx}{n}

%% file: section/1_intro.tex
\section{Introduction}
\label{sec:intro}

Video world models can generate visually compelling clips along camera trajectories \citep{huang2025voyager}. However, maintaining scene consistency over long rollouts remains challenging. When the camera returns to a previously observed region, its layout, appearance, and objects should remain consistent with earlier frames. As generation proceeds, these observations fall outside the generator's limited temporal context, making them difficult to recover from recent frames alone \citep{xiao2026worldmem,wu2026video}. Persistent scene memory is therefore important for consistent long-horizon generation \citep{zhao2026spatia}.

\begin{figure}[t]
\centering
\setlength{\abovecaptionskip}{4pt}
\makebox[1.0\textwidth][l]{%
\includegraphics[width=1.0\linewidth]{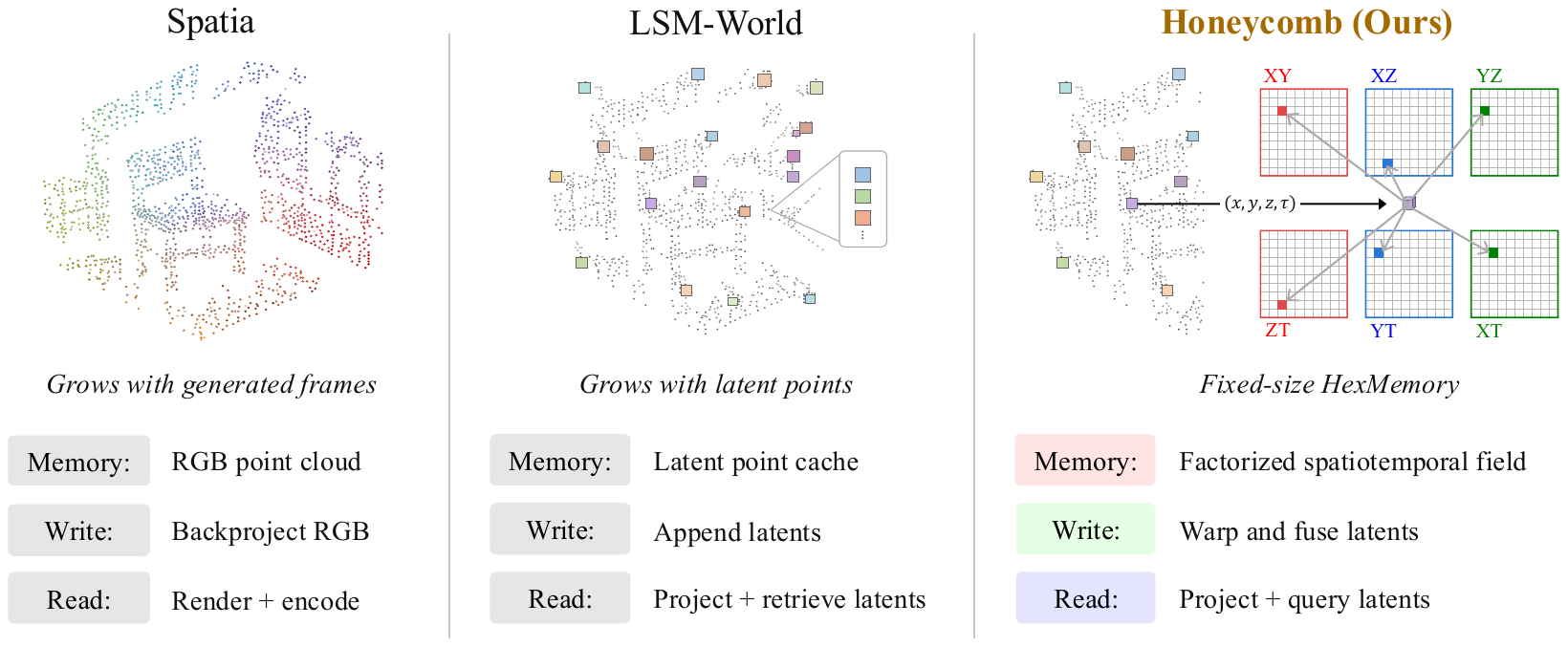}}
\vspace{-1.5em}
\caption{\textbf{Memory design comparison.} Spatia backprojects RGB observations into a point cloud, while LSM-World accumulates latent points. \method\ instead writes to HexMemory with fixed-size feature storage.}
\label{fig:memcompare}
\end{figure}

Existing methods address this challenge by storing observations in explicit spatial memories and projecting them into future views. Spatia maintains an updatable point cloud of RGB observations, while LSM-World stores diffusion latents associated with 3D points \citep{zhao2026spatia,wang2026latent}. These memories allow the generator to retrieve previously observed content, but their storage grows as new observations accumulate (Figure~\ref{fig:memcompare}). This raises a central question: \emph{can a video world model retain scene information without continually expanding its feature memory?} Doing so requires incorporating new observations into a fixed-size representation while preserving information needed for revisits.

We introduce \method, a video world model built on our proposed HexMemory. HexMemory represents scene features using a low-rank factorization into three spatial and three spatiotemporal planes \citep{cao2023hexplane}, whose dimensions remain fixed throughout generation. The planes are initialized from the input frame and updated after each generated chunk. A feed-forward writer maps the new chunk's latent observations into plane features, which are then fused with the existing memory. This recurrent update processes only the new observations, thus avoiding per-scene optimization and repeated processing of the entire video history. Before generating the next chunk, a memory reader retrieves latents from the planes to condition the video generator.

HexMemory consolidates multiple observations into shared plane features, while joint writer–reader training encourages these features to retain the latent information needed for accurate reconstruction. As illustrated in Figure~\ref{fig:teaser}, \method\ thus better preserves scene layout and appearance when the camera revisits earlier views, while the compared methods exhibit noticeable changes in geometry and texture.

Experiments on WorldScore and RealEstate10K demonstrate robust generation quality, novel-view synthesis, and revisit consistency. When evaluated without dynamic object filtering during memory writes, \method\ outperforms Spatia and LSM-World on overall WorldScore. For novel-view synthesis on RealEstate10K, \method\ achieves 18.45 dB PSNR, compared with 15.58 dB for Spatia and 17.46 dB for LSM-World. These results are achieved while keeping feature memory fixed in size. Our ablations show that a compact configuration uses 19.8 MB of feature storage (i.e., 73\% less), with only a 0.12 dB decrease in WorldScore closed-loop PSNR relative to the default. Recurrent writing achieves quality comparable to rebuilding memory from all observations while keeping write time constant.

Our contributions are threefold:
\begin{itemize}
    \item We introduce \method, a video world model with HexMemory, a fixed-size memory that stores persistent scene features in a low-rank representation.
    \item We develop a feed-forward writer that recurrently incorporates new observations into the memory without per-scene optimization or reprocessing the full history.
    \item We evaluate generation quality and revisit consistency on WorldScore and RealEstate10K, visualize the stored memory, and analyze the trade-offs between feature memory size, write cost, and quality.
\end{itemize}

% \FloatBarrier

%% file: section/2_related.tex
\section{Related Work}
\label{sec:related}

\subsection{Camera Conditioning for Video Generation}

Camera trajectories provide a controllable mechanism for video generation. Early approaches inject explicit camera poses or motion representations through auxiliary control modules~\citep{wang2024motionctrl,he2025cameractrl}. 
As video generation backbones transition toward large diffusion transformers, recent work has investigated architectural and training strategies for camera conditioning~\citep{bahmani2025ac3d,bahmani2025vd3d}. 
To improve cross-view consistency, other work has incorporated geometric priors such as epipolar constraints~\citep{xu2024camco,zheng2024cami2v} and richer camera representations, including relative positional encodings and geometry-aware tokens~\citep{zhang2026unified,li2026rerope,zhao2026cetcam}.
Related approaches ground view synthesis and camera-controlled generation in explicit scene geometry through point-based representations, reconstructed 3D proxies, target-view reprojections, or anchor views~\citep{zhou2023dynpoint,yu2024viewcrafter,li2025realcam,ren2025gen3c,wang2026epic}. 
As video world models expand to iterative camera exploration, autoregressive streaming, and video-action modeling~\citep{he2025cameractrlii,zhao2026geostream,liu2026driveva,liu2026universe, zhang2026world}, persistent memory becomes increasingly useful for maintaining scene consistency over long rollouts. This motivates persistent memory mechanisms that store, update, and recover scene information outside the model's temporal context.

\subsection{Persistent Memory for Long-Horizon Video Generation}

Dedicated benchmarks assess scene recovery across long temporal or viewpoint gaps~\citep{lian2025loopnav,ye2026mind,zhang2026mbench,wu2026addressable} and whether unobserved processes evolve consistently~\citep{ma2026out,duan2026liveworld,lu2026current,chen2026memobench}.
At the model level, causal autoregressive diffusion and history conditioning reuse recent frames or latents across rollout steps~\citep{
chen2024diffusion,song2025history}. 
To extend this context to longer histories, observation- and context-based memories retain, compress, or retrieve past observations~\citep{
yu2025context,li2025vmem,oshima2025worldpack,xiao2026worldmem,guo2026memorize,
zhou2026stream3d,yu2026memlearner,xue2026ring,xu2026wonder,
yi2026worldkv,zhang2025frame}.
Within this family, long-range history is represented through latent tokens, recurrent states, positional states,
or attention-space memories, including motion-aware retrieval for dynamic subjects~\citep{
yu2025videossm,xu2026ucm,yang2026decmem,wu2026addressable,chen2026out}.
The storage cost of retaining such histories has motivated fixed-size latent representations, bounded key--value caches, and pose-indexed history banks with limited capacity~\citep{wei2026geometry,kim2026memrope,chen2026reworld,wu2026infinite}.

Complementary to these approaches, spatial memories associate stored scene content with explicit 3D structure. Early scene-expansion systems accumulated geometric representations for interactive exploration~\citep{yu2024wonderjourney,yu2025wonderworld}, while recent methods update persistent scene memories during autoregressive rollouts~\citep{
wang2025evoworld,wu2026video,zhao2026spatia,yu2026mosaicmem,lee20263d,wang2026anchorweave,garcin2026persist}.
Spatia~\citep{zhao2026spatia} and LSM-World~\citep{wang2026latent} cache RGB observations and diffusion latents respectively, so their memory footprint grows as observations accumulate. To address this growth, HexMemory stores scene features in a fixed-size low-rank representation, drawing on plane factorizations~\citep{cao2023hexplane,fridovich2023k}. We integrate this memory into \method, a video world model designed to preserve scene consistency over rollouts while keeping feature storage constant.

%% file: section/3_method.tex
\section{Method}
\label{sec:method}

\subsection{Overview}
\label{sec:overview}

\vspace{-0.5em}
\begin{figure}[H]
\centering
\setlength{\abovecaptionskip}{4pt}
\makebox[1.0\textwidth][l]{%
\includegraphics[width=1.0\linewidth]{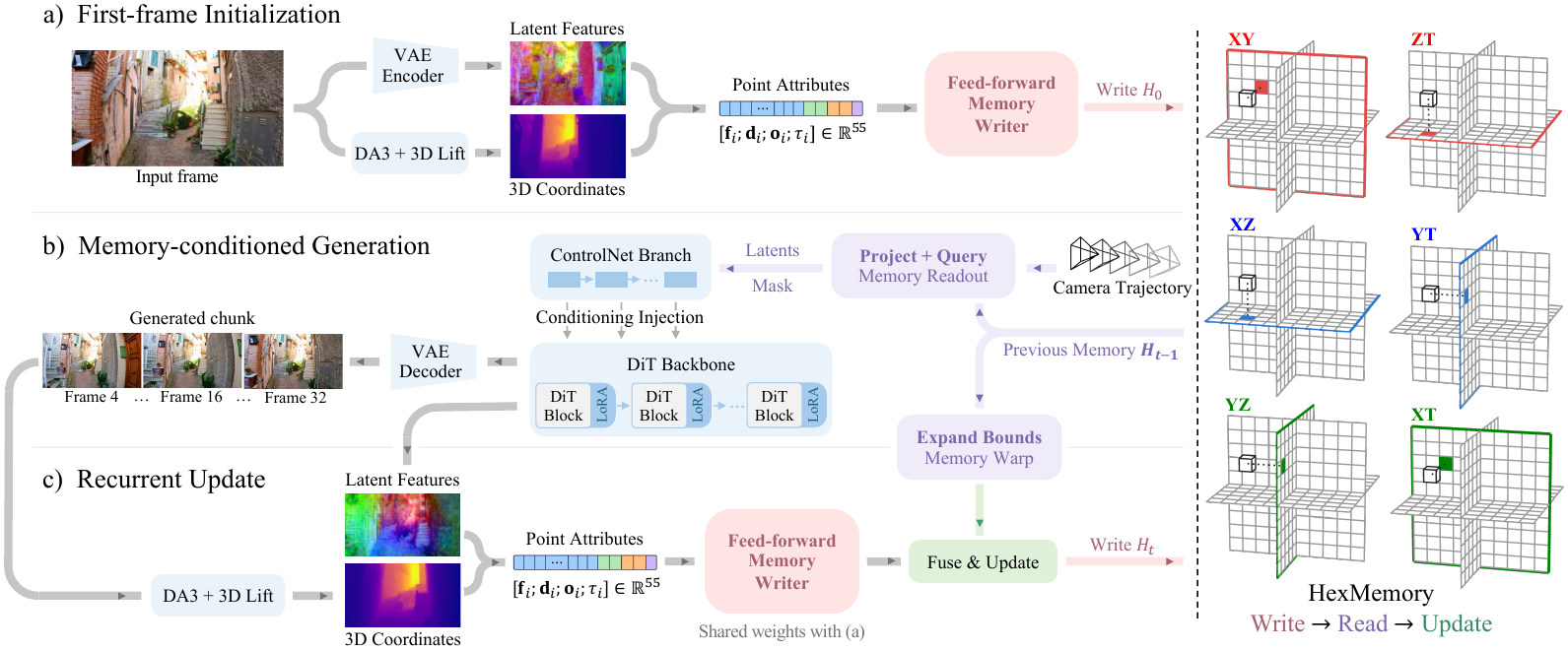}}
\vspace{-1em}
\caption{\textbf{Overview of the \method\ pipeline.} For scene-consistent video generation, we a) initialize a fixed-size HexMemory, b) read it to condition a generated chunk, and then c) write new observations back into the same tensors.}
\label{fig:method-overview}
\end{figure}

\method\ maintains scene memory in a HexMemory $\hexmem$ of six fixed-size feature planes. We initialize the memory from the input frame, then read it to condition each new chunk and write it back into memory (Figure~\ref{fig:method-overview}). Unlike LSM-World, which appends latent points to a growing cache, \method\ writes into the existing planes,
\begin{equation}
\hexmem_t = U_\theta\!\left(\hexmem_{t-1}, W_\theta(o_t)\right),
\label{eq:recurrent-state}
\end{equation}
where $o_t$ contains the chunk's latent points and observation attributes, $W_\theta$ is a feed-forward writer, and $U_\theta$ fuses its output with the previous memory.
\textbf{Initialization (Figure~\ref{fig:method-overview}~(a)).} We encode the input frame into latent features and lift them into world space using estimated depth. The writer maps these latent points into three spatial and three spatiotemporal planes (Sections~\ref{sec:state} and~\ref{sec:writer}).
\textbf{Readout and generation (Figure~\ref{fig:method-overview}~(b)).}
% A geometry index $\visindex$ records point positions and write times. 
We project the 3D points into the target views to determine visibility, then query $\hexmem$ at the selected positions and write times. The reconstructed latents and visibility masks condition the diffusion transformer (DiT) through a side branch (Sections~\ref{sec:readout} and~\ref{sec:generation}).
\textbf{Memory write-back (Figure~\ref{fig:method-overview}~(c)).} After generation, we estimate depth from the decoded frames and lift the generated latents into world space. Unlike Spatia and LSM-World, we do not exclude dynamic objects or sky from memory writes. We warp the previous planes when the spatial or temporal bounds expand, then fuse them with the writer's output for the new chunk (Section~\ref{sec:update}).

\subsection{HexMemory}
\label{sec:state}

We represent the memory as six 2D feature planes, one for each pair of the four coordinate axes, namely the three spatial axes and the write time $\tau$. Following HexPlane \citep{cao2023hexplane}, the planes form three pairs with orthogonal axes,
\begin{equation}
\hexmem=\big\{(S_{XY},T_{ZT}),\;(S_{XZ},T_{YT}),\;(S_{YZ},T_{XT})\big\},
\label{eq:state}
\end{equation}
where $S$ and $T$ denote spatial and spatiotemporal planes, respectively. Each pair covers all four coordinates exactly once. Every plane is a 2D grid of $R$-dimensional features. The memory also stores a bounding box $B$ over world space and write time, used to normalize coordinates to $[-1,1]$. Each plane carries a confidence map $N$ recording the accumulated interpolation weight at each cell.

At a world point $\vp=(x,y,z)$ and write time $\tau$, we extract the memory feature
\begin{align}
\phi(\vp,\tau)=\big[&S_{XY}(x,y)\odot T_{ZT}(z,\tau);\nonumber\\[-2pt]
&S_{XZ}(x,z)\odot T_{YT}(y,\tau);\nonumber\\[-2pt]
&S_{YZ}(y,z)\odot T_{XT}(x,\tau)\big]\in\RR^{3R},
\label{eq:hexread}
\end{align}
where each plane is sampled by bilinear interpolation at coordinates normalized to the current bounds. Spatial planes share information across write times, while their products with spatiotemporal planes capture time-varying content. The original HexPlane optimizes its planes separately for each scene. HexMemory instead uses a writer shared across scenes to construct and recurrently update the memory without per-scene optimization.

% \FloatBarrier
\begin{figure}[t]
\centering
\setlength{\abovecaptionskip}{4pt}
\makebox[0.7\textwidth][l]{%
\includegraphics[width=0.7\linewidth]{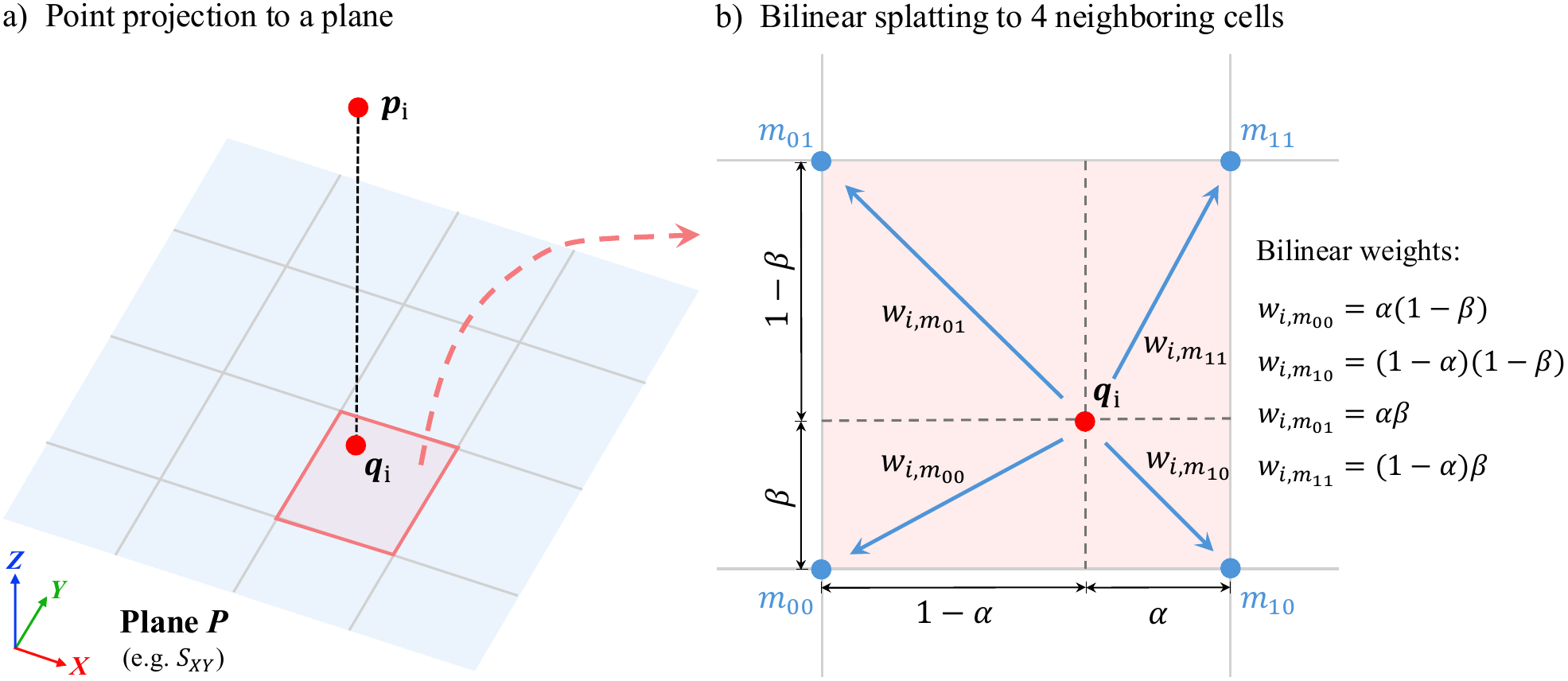}}
\caption{\textbf{Bilinear splatting within the feed-forward writer.}
(a) Point $p_i$ maps to location $q_i$ on plane $P$ and (b) its learned contribution $\mathbf{c}_i^P$ is distributed to four neighboring cells using bilinear weights $w_{im}$. This operation is repeated for all six planes.}
\label{fig:bilinear-splatting}
\end{figure}

\subsection{Feed-Forward Memory Writer}
\label{sec:writer}

The writer converts each chunk’s latent observations into plane features. Using depth and camera parameters, we backproject each valid latent grid location into a 3D world-space position and pair it with the corresponding latent token $\mathbf{f}_i$. Each point also carries the camera center $\vo_i$ and ray direction $\vd_i$.

A small network maps each point's attributes to one contribution $\vc_i^{P}$ per plane $P$. Here, $p_i=(x_i,y_i,z_i)$ is the normalized 3D position
of point $i$ and $\tau_i$ is its normalized write time. The contribution is then placed on that plane at the point's two coordinates, for example at $(x_i,y_i)$ on $S_{XY}$ and at $(z_i,\tau_i)$ on $T_{ZT}$, and is distributed over the four surrounding grid cells with bilinear weights, as shown in Figure~\ref{fig:bilinear-splatting}. Each cell averages what it receives,
\begin{equation}
\bar{\mathbf{C}}^{P}_m=\frac{\sum_i w_{im}\,\vc_i^{P}}{\sum_i w_{im}},\qquad
N_m=\sum_i w_{im},
\label{eq:splat}
\end{equation}
where $w_{im}$ is the bilinear weight of point $i$ for cell $m$ and $N_m$ is the total weight received. Cells with no contributions are zero-filled. Lightweight networks map the averaged contributions and accumulated weights to the six feature planes. The weights also serve as confidence maps for subsequent writes.

We jointly train the writer, fusion networks (Section~\ref{sec:update}), and reader (Section~\ref{sec:readout})
with a latent reconstruction objective. Each training clip provides
latents, estimated depth, and camera poses. We write the input frame
followed by two chunks, matching the rollout schedule. After each
write, we query the memory from the clip's camera views and minimize
mean squared error between the reconstructed latents and the original
tokens associated with the visible points selected by projection.
The loss is computed in normalized latent space.

\subsection{Recurrent Memory Update}
\label{sec:update}

For each subsequent chunk, the writer processes only the
new latent points and produces planes to fuse with the previous memory $\hexmem_{t-1}$ (Figure~\ref{fig:method-overview}~(c)). Before writing, we expand the bounds from $B_{t-1}$ to $B_t$ as needed to include the new observations, extending the time range in whole-chunk increments. We warp the previous planes and their confidence maps into $B_t$ by bilinear interpolation at corresponding world coordinates, preserving their grid dimensions. Newly covered regions receive zero confidence, with spatial planes filled with zero and spatiotemporal planes with one. As the bounds expand, the fixed grids represent larger spatial and temporal ranges at coarser resolution.

Let $P^{o}$ be a warped previous plane with confidence $N^{o}$, and $P^{n}$ the writer's new plane for the chunk with confidence $N^{n}$ from Eq.~\ref{eq:splat}. The two are pooled cell by cell, weighting each by its confidence,
\begin{equation}
\bar P=\frac{N^{o}P^{o}+N^{n}P^{n}}{N^{o}+N^{n}},
\label{eq:pool}
\end{equation}
and for a cell with zero total confidence, the pooled value is set to the previous value. A small network per plane then corrects the pooled result,
\begin{equation}
P_t=\bar P+h_P\big(P^{o},P^{n},\bar P,N^{o},N^{n}\big),
\qquad
N_t=N^{o}+N^{n},
\label{eq:fuse}
\end{equation}
where the output layer of $h_P$ is zero-initialized, so fusion begins as confidence-weighted pooling and learns a residual correction.

\subsection{Memory Readout}
\label{sec:readout}

To provide memory conditioning for video generation, we reconstruct a latent feature map from HexMemory for the target camera view. We first project the 3D points into the target view at latent resolution. For each latent cell, we retain the nearest projected point in front of the camera. A visibility mask $m^t$ marks cells containing a point.

For each occupied cell $(u,v)$, let $i$ denote the selected point. We query HexMemory at its world position $\vp_i$ and source write time $\tau_i$, then decode the resulting feature with the shared reader:
\begin{equation}
\hat{\vz}^t(u,v) = g\bigl(\phi(\vp_i,\tau_i),\,\vd^t_{uv},\,\vo^t\bigr).
\label{eq:projection}
\end{equation}
Here, $\phi$ is the memory feature defined in Eq.~\ref{eq:hexread}, while $\vd^t_{uv}$ and $\vo^t$ are the target-view ray direction and normalized camera center. Cells receiving no point are filled with zeros. The reconstructed latent map $\hat{\vz}^t$ and visibility mask $m^t$ are then supplied to the DiT as memory conditioning.

\subsection{Video Generation and Memory Write-Back}
\label{sec:generation}

We build on a pretrained camera-controllable video diffusion model and generate video autoregressively in overlapping chunks. Each chunk is denoised from noise while keeping its first latent fixed to the encoded input frame or the shared boundary latent from the preceding chunk. The reconstructed latents $\hat{\vz}^t$ and mask $m^t$ enter the denoiser through a ControlNet-style branch \citep{jiang2025vace}, as shown in Figure~\ref{fig:method-overview}~(b).

After a chunk is denoised, its frames are decoded and a monocular depth estimator provides depth for frames corresponding to latent timesteps at their trajectory poses. The chunk's new latent frames are then written into the
memory as described in Section~\ref{sec:update}, each at its own write time (Figure~\ref{fig:method-overview}~(c)).

We adapt the generator using precomputed readouts from the frozen memory model. Following Spatia, we apply noise augmentation to preceding-frame latents to mitigate the distribution mismatch between ground-truth conditioning during training and generated conditioning during inference.

\begin{figure}[t]
\centering
\makebox[1.0\textwidth][l]{%
\includegraphics[width=1.0\linewidth]{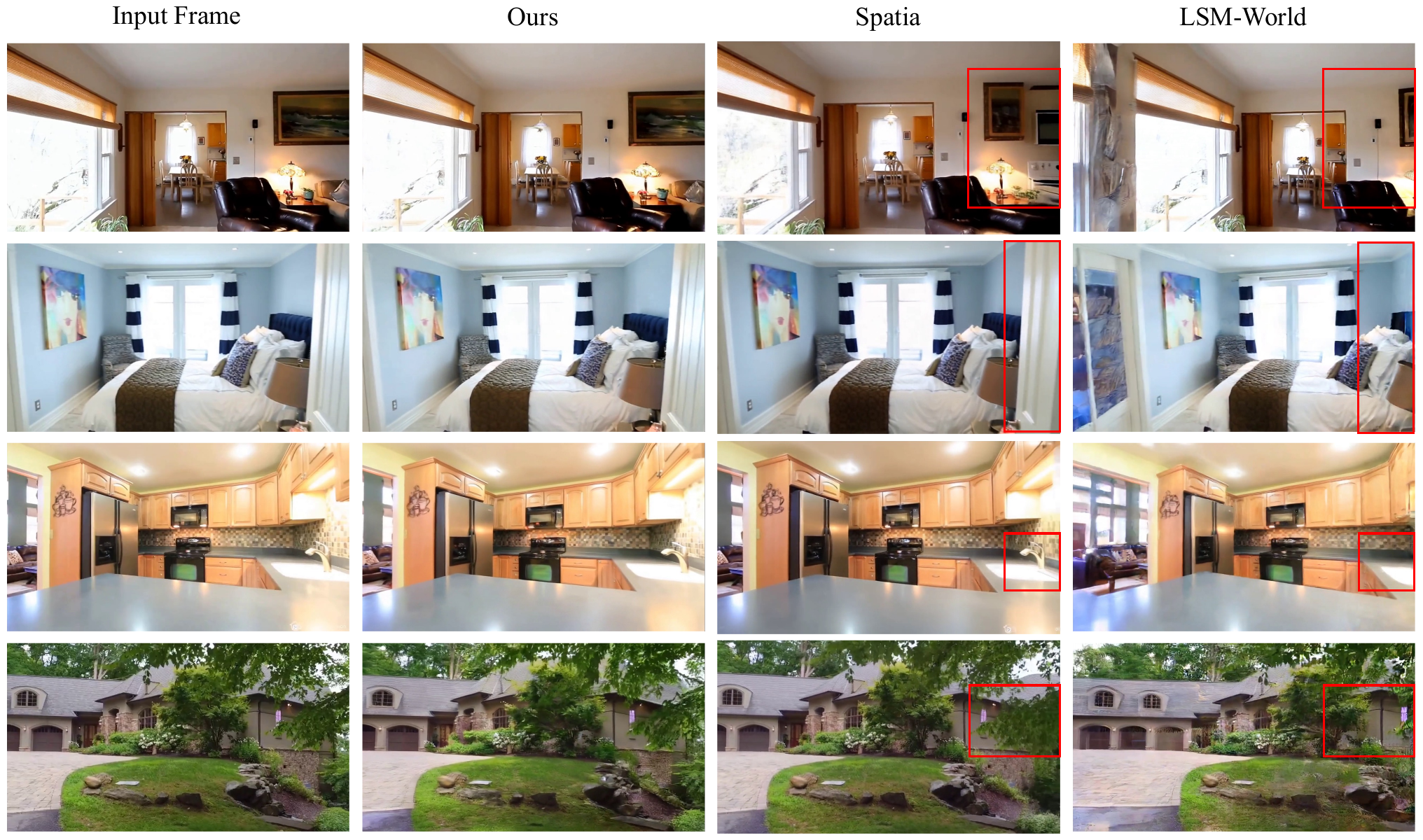}}
\vspace{-1.5em}
\caption{\textbf{Closed-loop comparison.} \method\ preserves object appearance and scene layout, while Spatia and LSM-World show changes in furnishings, geometry, and texture.}
\label{fig:in-domain-layout}
\end{figure}

%% file: section/4_exp.tex
\section{Experiments}
\label{sec:experiments}

\subsection{Training and Inference Setup}
\label{sec:training}
Our backbone is Wan2.2 \citep{wan2025} with 5B parameters. The ControlNet branch receives 48 reconstructed latent channels and a visibility mask. Its eight blocks connect to every fourth backbone block starting from the first and are initialized from the corresponding backbone blocks. We train on RealEstate10K \citep{zhou2018stereo}, with camera poses, intrinsics, and depth estimated using ViPE \citep{huang2025vipe} configured with Depth Anything 3, and the same depth estimator provides depth for memory write-back at inference. The branch is first trained for 10,000 iterations with the backbone frozen. The branch is then frozen and the backbone is fine-tuned with LoRA \citep{hu2022lora} of rank~64 on the attention and feed-forward layers for 5,000 iterations. Both stages use AdamW with learning rates of $10^{-5}$ and $10^{-4}$ respectively and a total effective batch size of 64 on H200 GPUs. Each generation chunk contains nine latent frames at $44\times80$, corresponding to 33 RGB frames at $704\times1280$. At inference we use the UniPC scheduler with 40 sampling steps. The memory planes have rank $R=48$.

\subsection{Evaluation and Main Results}
We evaluate \method\ on generation quality and memory effectiveness. For generation quality, we evaluate it without dynamic object filtering on WorldScore \citep{duan2025worldscore}, which contains 3,000 image-to-video samples that evaluate static and dynamic aspects of world generation. We also evaluate on 100 videos from the RealEstate10K test set, conditioning on the first frame and following the ground-truth camera trajectory. We report PSNR, SSIM, and LPIPS against the original frames. For memory effectiveness, we follow the closed-loop evaluation setting of Spatia \citep{zhao2026spatia}. Starting from the input images of 100 WorldScore scenes, we generate videos along camera trajectories that leave and return to the initial viewpoint. We compare the final frame with the input image using the same metrics. We additionally report flow error, the RAFT \citep{teed2020raft} optical-flow magnitude in pixels between the input image and the final frame.

As shown in Table~\ref{tab:results}, \method\ achieves the best novel-view synthesis results among the evaluated methods across PSNR, SSIM, and LPIPS. On WorldScore closed-loop evaluation, it also achieves the best results across all metrics, improving PSNR by 1.23\,dB over the next-best method and reducing flow error from 6.64 to 3.00 pixels relative to Spatia. These results demonstrate strong revisit fidelity alongside fixed-size HexMemory storage.

\begin{table}[t]
\caption{\textbf{Evaluation results on WorldScore.} The Average Score is the mean of the Static and Dynamic Scores; all remaining metrics are computed by the WorldScore benchmark.}
\label{tab:world-score}
\centering
\small
\setlength{\tabcolsep}{4pt}
\begin{tabular}{@{}lccc|cccc@{}}
\toprule
\textbf{Method}
& \makecell{Average\\Score} & \makecell{Static\\Score} & \makecell{Dynamic\\Score}
& \makecell{3D\\Const} & \makecell{Photo\\Const}
& \makecell{Style\\Const} & \makecell{Subject\\Quality}\\
\midrule
\multicolumn{8}{@{}l}{\textit{Models with 3D cache}}\\
WonderJourney & 54.19 & 63.75 & 44.63 & 80.60 & 79.03 & 62.82 & \textbf{66.56}\\
WonderWorld   & 61.79 & \textbf{72.69} & 50.88 & \textbf{86.87} & 85.56 & 70.57 & 49.81\\
Spatia        & \underline{63.21} & 64.88 & \underline{61.54} & \underline{83.26} & \textbf{89.09} & \underline{83.33} & 46.66\\
LSM-World     & 61.20 & 62.69 & 59.70 & 80.88 & 76.10 & -- & --\\
\midrule
\multicolumn{8}{@{}l}{\textit{General video models}}\\
VideoCrafter2 & 50.03 & 52.57 & 47.49 & 65.14 & 61.85 & 43.79 & 56.74\\
EasyAnimate   & 52.25 & 52.85 & 51.65 & 67.29 & 47.35 & 73.05 & 50.31\\
Allegro       & 53.64 & 55.31 & 51.97 & 70.50 & 69.89 & 65.60 & 47.41\\
Wan2.1        & 55.21 & 57.56 & 52.85 & 78.74 & 78.36 & 77.18 & \underline{59.38}\\
\midrule
\method & \textbf{65.52} & \underline{68.01} & \textbf{63.03} & 82.29 & \underline{85.76} & \textbf{84.21} & 46.28\\
\bottomrule
\end{tabular}
\vspace{-0.5em}
\end{table}

\begin{table}[h]
\caption{\textbf{Novel-view synthesis on RealEstate10K and closed-loop on WorldScore.} We report evaluation results of all baseline methods using their default settings.}
\label{tab:results}
\centering
\small
\setlength{\tabcolsep}{4pt}
\begin{tabular}{@{}lccc|cccc@{}}
\toprule
& \multicolumn{3}{c}{RE10K NVS} & \multicolumn{4}{c}{WorldScore closed-loop}\\
\cmidrule(lr){2-4}\cmidrule(l){5-8}
Method & PSNR$\uparrow$ & SSIM$\uparrow$ & LPIPS$\downarrow$
& PSNR$_C$$\uparrow$ & SSIM$_C$$\uparrow$ & LPIPS$_C$$\downarrow$ & Flow$_C$$\downarrow$\\
\midrule
ViewCrafter & 12.28 & 0.512 & 0.571 & 12.32 & 0.369 & 0.574 & 30.78 \\
FlexWorld   & 13.17 & 0.567 & 0.544 & 12.86 & 0.430 & 0.602 & 55.77 \\
Voyager     & 14.67 & 0.577 & 0.493 & \underline{15.99} & 0.459 & 0.423 & 7.11 \\
Spatia      & 15.58 & 0.616 & \underline{0.390} & 15.67 & \underline{0.488} & \underline{0.353} & \underline{6.64}\\
LSM-World   & \underline{17.46} & \underline{0.636} & 0.452 & 15.12 & 0.460 & 0.463 & 27.05\\
\method\    & \textbf{18.45} & \textbf{0.674} & \textbf{0.274} & \textbf{17.22} & \textbf{0.504} & \textbf{0.311} & \textbf{3.00}\\
\bottomrule
\end{tabular}
\vspace{-0.5em}
\end{table}

\begin{figure}[t]
\centering
\makebox[1.0\textwidth][l]{%
\includegraphics[width=1.0\linewidth]{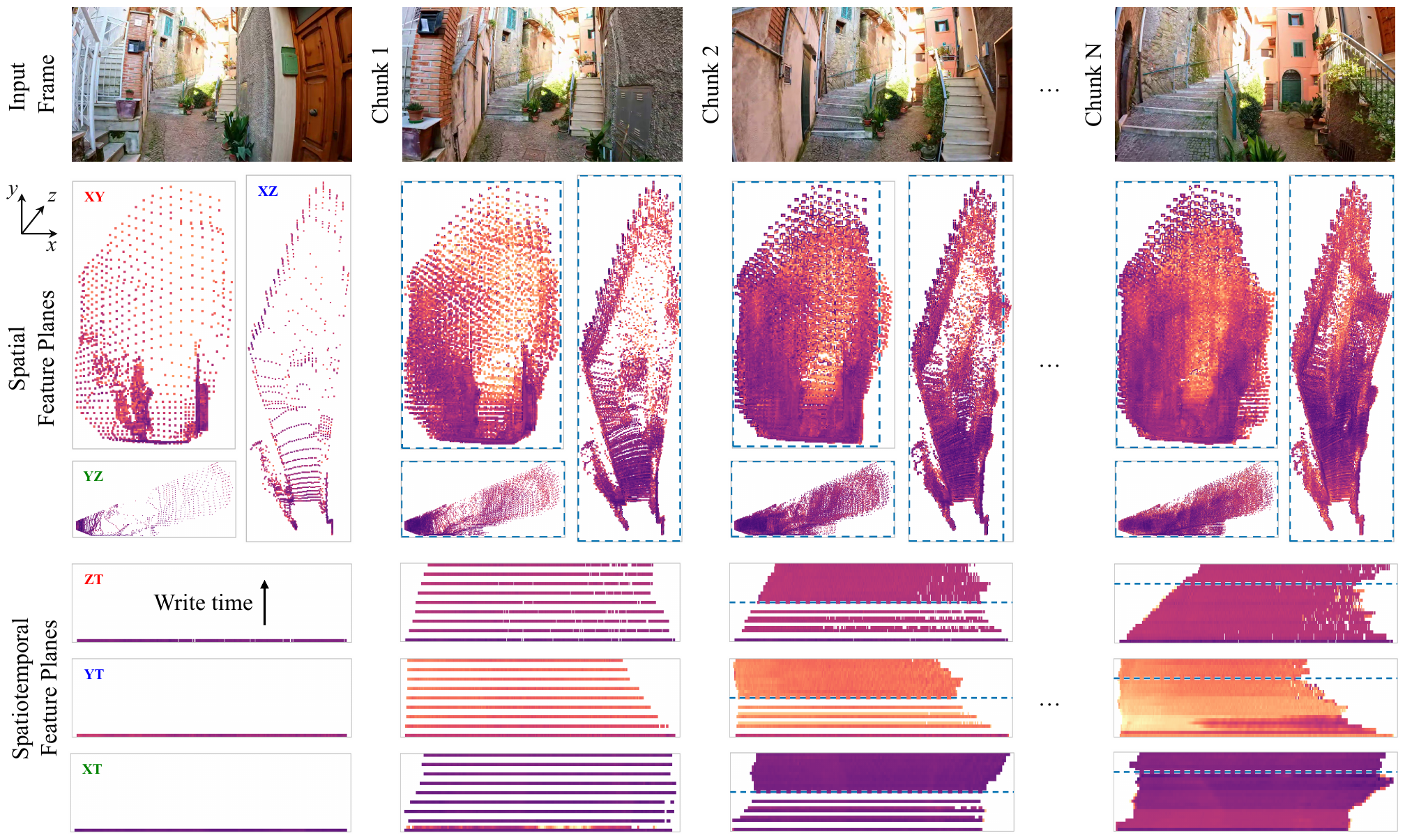}}
\vspace{-1em}
\caption{\textbf{Visualization of HexMemory.} Three spatial and three spatiotemporal planes are shown at initialization and after successive chunks. Dashed boundaries mark the extent of the previous memory after warping into the updated coordinate range.}
\label{fig:hex-vis}
\end{figure}

\subsection{HexMemory Visualization}
We visualize the spatial feature planes from initialization through memory updates (Figure~\ref{fig:hex-vis}). Each cell is colored by the $\ell_2$ norm of its feature vector, with cells of negligible confidence left blank. The three spatial planes $XY$, $XZ$, and $YZ$ provide front, top-down, and side views of the alley, respectively. The corridor layout and rising staircase are visible in the feature maps. As new regions are observed, the previous planes are warped to accommodate the expanded spatial coverage, then fused with features from the new chunk. Previously observed structure remains visible as additional regions are incorporated. 

By organizing observations across spatial and spatiotemporal feature planes, HexMemory retains scene information that can be retrieved at corresponding world-space locations when generating new views. The retrieved features condition subsequent generation, helping preserve scene layout and appearance during both novel-view synthesis and revisits. As the spatiotemporal planes are less intuitive to interpret, we explain them more in the Appendix.

\subsection{Ablation Studies}

% \vspace{-1em}
\begin{table}[!ht]
\caption{\textbf{HexMemory resolution ablation.}
We vary the plane resolution and evaluate on WorldScore closed-loop samples.}
\label{tab:ablation-size}
\centering
\small
\begin{tabular}{@{}lc|ccc@{}}
\toprule
Resolution & HexMemory (MB)$\downarrow$
& PSNR$_C$$\uparrow$ & SSIM$_C$$\uparrow$
& LPIPS$_C$$\downarrow$\\
\midrule
512 & 73.9 & 17.22 & 0.504 & 0.311\\
384 & 42.6 & 17.14 & 0.503 & 0.313\\
256 & 19.8 & 17.10 & 0.500 & 0.319\\
128 &  5.8 & 16.77 & 0.490 & 0.338\\
\bottomrule
\end{tabular}
\vspace{-1em}
\end{table}

\begin{table}[!ht]
\caption{\textbf{Memory writer ablation.}
We hold the plane resolution fixed and compare write time as well as the WorldScore closed-loop performance.}
\label{tab:ablation-writer}
\centering
\small
\setlength{\tabcolsep}{4pt}
\begin{tabular}{@{}lccc|ccc@{}}
\toprule
& \multicolumn{3}{c}{Write time (ms)$\downarrow$}
& \multicolumn{3}{c}{WorldScore closed-loop}\\
\cmidrule(lr){2-4}\cmidrule(lr){5-7}
Writer & Chunk 2 & Chunk 5 & Chunk 9
& PSNR$_C$$\uparrow$ & SSIM$_C$$\uparrow$
& LPIPS$_C$$\downarrow$\\
\midrule
Direct optimization & 3217 & 3228 & 3130 & 17.44 & 0.510 & 0.299\\
Replacement         & 11.2 & 23.3 & 40.1 & 17.16 & 0.502 & 0.313\\
Recurrent           & 13.1 & 13.2 & 13.2 & 17.22 & 0.504 & 0.311\\
\bottomrule
\end{tabular}
% \vspace{-0.5em}
\end{table}

We evaluate the HexMemory resolution and memory writer on the WorldScore closed-loop samples. First, the resolution of the feature planes determines the size of $\hexmem$. We vary it while holding all other settings fixed. As shown in Table~\ref{tab:ablation-size}, lowering the resolution substantially reduces memory with little effect on revisit consistency, while further reductions lead to more noticeable degradation. This suggests that smaller planes can retain much of the scene information needed for consistent revisits.

Next, we compare three ways to write $\hexmem$: per-scene direct optimization, which fits the planes by gradient descent as in the original HexPlane method; a replacement writer, which rebuilds the planes from all stored points at each chunk; and our recurrent writer, which processes only the new chunk and fuses its features with the existing memory. As shown in Table~\ref{tab:ablation-writer}, the replacement writer achieves similar quality to ours, but its write cost grows with the rollout as it reprocesses the entire history, whereas ours remains constant at 13~ms. Direct optimization attains slightly higher quality but is roughly 240× slower per write, further highlighting the efficiency of feed-forward writing.

%% file: section/5_conclusion.tex
\section{Conclusion}
\label{sec:conclusion}

Maintaining scene consistency over long video rollouts requires persistent memory that can efficiently incorporate new observations. In this work, we introduce \method, a video world model with HexMemory. This design keeps feature storage fixed without reprocessing the entire history at each write. Experiments on WorldScore and RealEstate10K demonstrate better generation quality, novel-view synthesis, and revisit consistency. Ablation studies show that recurrent writing maintains quality with constant write cost. These results highlight the potential of recurrent feature memory for efficient and consistent video world modeling.

%% file: section/6_appendix.tex
\appendix
\section*{Appendix}

\section{Additional Implementation Details}
\label{app:checklist}

% \paragraph{Reference-frame selection.}
% For autoregressive continuation, each chunk uses 33 target RGB frames $T$, eight preceding RGB frames $P$, and up to eight reference frames $R$. Their latents are arranged as $[R \mid P \mid T]$, combining earlier visual observations with recent temporal context. Reference candidates are drawn from frames before the preceding window. During training, we compute each frame's voxel occupancy by back-projecting its depth into world coordinates using its camera parameters, with a voxel size of $0.1$. Each candidate is scored by its maximum occupancy intersection-over-union (IoU) with an individual target frame. During inference, we use the stored geometry to obtain voxel occupancy within the camera views and score each historical candidate against the union of occupancy sets across the target views. We retain candidates with scores of at least $0.04$ and select the eight highest-scoring frames, or all qualifying frames when fewer are available.

\paragraph{Preceding-latent augmentation.}
To support better autoregressive generation, we apply low-level noise augmentation to the preceding-frame latents during stage-two training, following Spatia~\citep{zhao2026spatia}. Here, $P$ denotes the eight RGB frames immediately preceding the shared boundary frame, and $z_P$ denotes their jointly encoded video latents. For each training sample, we draw $t_{\mathrm{aug}} \sim \mathcal{U}(0,50)$ and form
\begin{equation}
    \widetilde{z}_P
    = (1-\sigma_{\mathrm{aug}})z_P
      + \sigma_{\mathrm{aug}}\epsilon,
    \qquad
    \sigma_{\mathrm{aug}}=\frac{t_{\mathrm{aug}}}{1000},
    \quad
    \epsilon\sim\mathcal{N}(0,I).
\end{equation}
The sampled noise level is shared across the sample's preceding latent frames. The augmented latents retain timestep-zero conditioning in the denoiser, while the flow-matching objective~\citep{lipman2023flow} supervises the target latents after the first-frame anchor. Stage one uses clean preceding latents.

\paragraph{Latent re-encoding during rollout.}
The first chunk is conditioned on the input image using the pretrained backbone and the stage-one conditioning branch. From the second chunk onward, we activate the stage-two LoRA adapters and prepare conditioning inputs from generated RGB frames obtained through joint decoding of the accumulated latent sequence. The eight RGB frames immediately before the shared boundary frame are encoded together as a video, producing two preceding latents. The boundary RGB frame is also encoded independently and supplies the first latent of the next target chunk, which remains fixed throughout denoising. These encoding procedures follow the construction of the corresponding training inputs i.e., video encoding for preceding context and single-frame encoding for the anchor. For memory write-back, the writer receives the newly generated chunk latents together with geometry estimated from their corresponding decoded frames.

%% file: section/7_qualitative_figures.tex
% \clearpage
\section{Additional Qualitative Results}
\label{sec:qualitative-results-appendix}

We present extended closed-loop comparisons on WorldScore and RealEstate10K, followed by novel-view synthesis examples on RealEstate10K.

\begin{figure}[p]
\centering
\includegraphics[width=\linewidth]{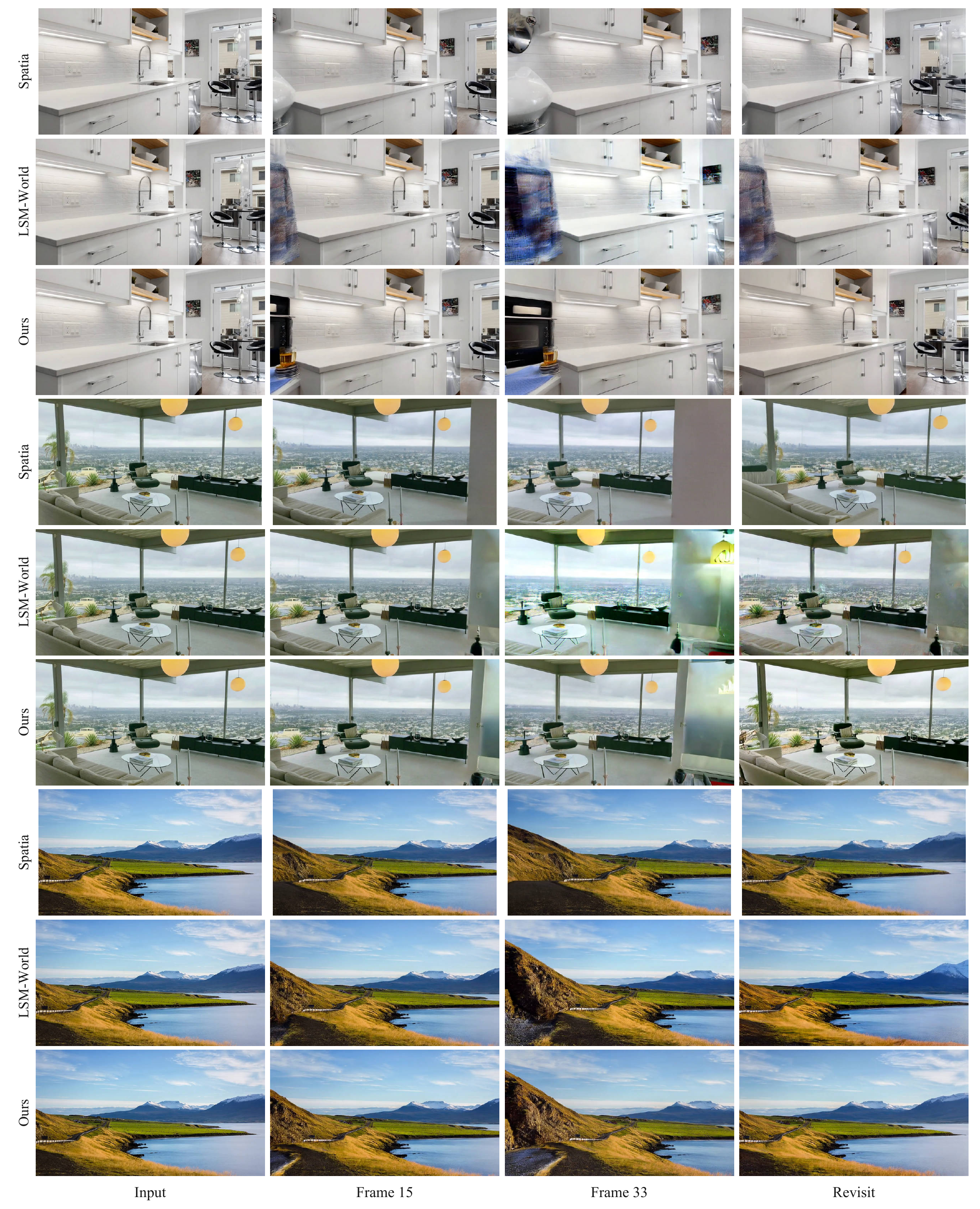}
\caption{\textbf{Extended closed-loop comparisons on WorldScore.}
For each of the three examples, rows show Spatia, LSM-World, and
\method, respectively.}
\label{fig:revisit-additional-1}
\end{figure}

\begin{figure}[p]
\centering
\includegraphics[
    width=\textwidth,
    height=0.92\textheight,
    keepaspectratio
]{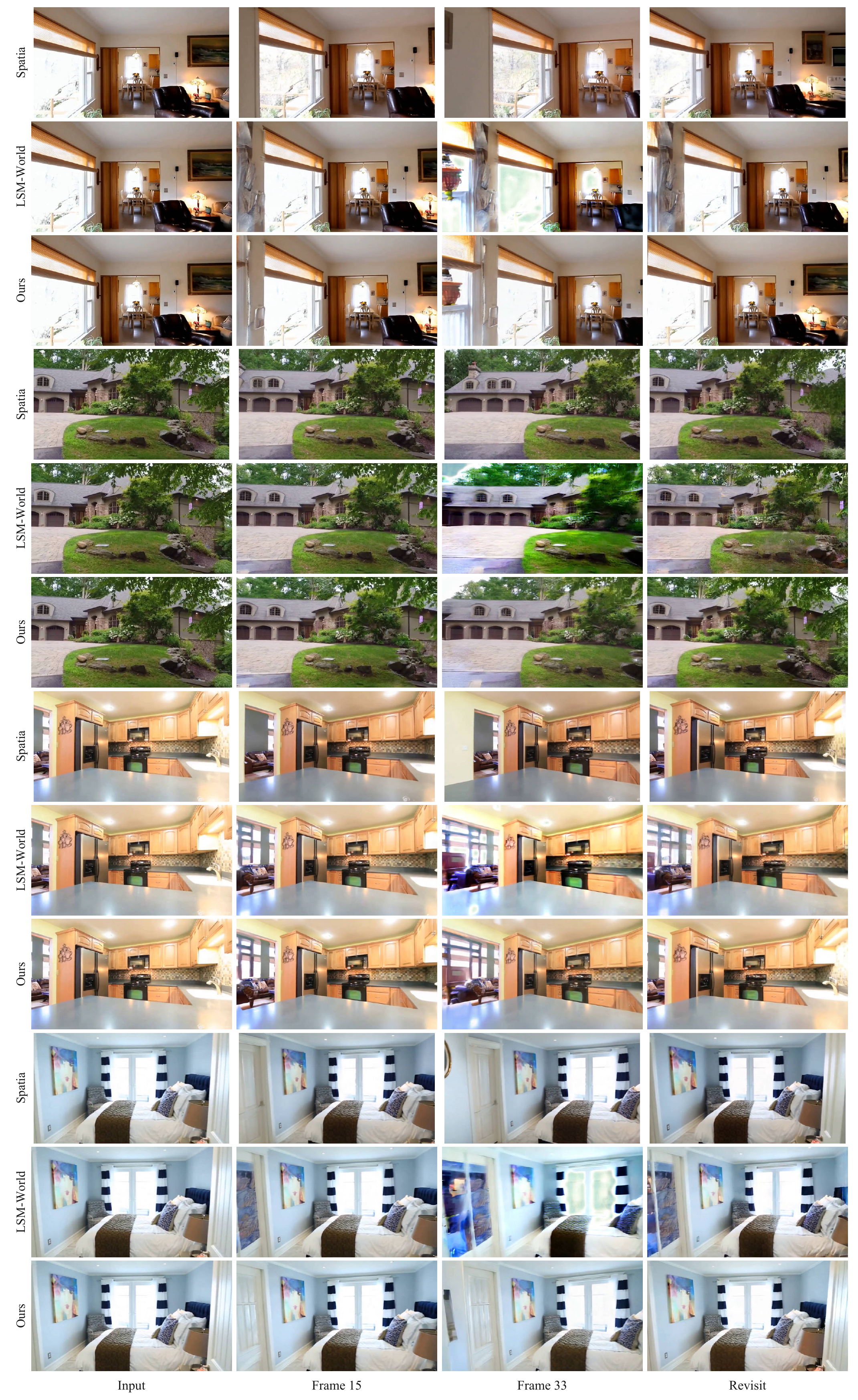}
\caption{\textbf{Extended closed-loop comparisons on RealEstate10K.}
For each of the four examples, rows show Spatia, LSM-World, and
\method, respectively.}
\label{fig:revisit-additional-2}
\end{figure}

\paragraph{Closed-loop generation.}
Figures~\ref{fig:revisit-additional-1} and~\ref{fig:revisit-additional-2}
extend the seven examples presented in the main manuscript, grouped into
three WorldScore examples and four RealEstate10K examples. For each example,
we show the input image, frames 15 and 33, and the final revisit frame.
The intermediate views provide context for the camera's movement through
the scene, while the final frame allows comparison with the initial
observation.

In the illustrated examples, Spatia and LSM-World exhibit appearance changes and artifacts in intermediate views, while \method\ better preserves scene appearance along the trajectory and recovers furnishings and scene structures at the final revisit.

\clearpage

\clearpage

\clearpage

\begin{figure}[!htbp]
\centering
\includegraphics[width=\linewidth]{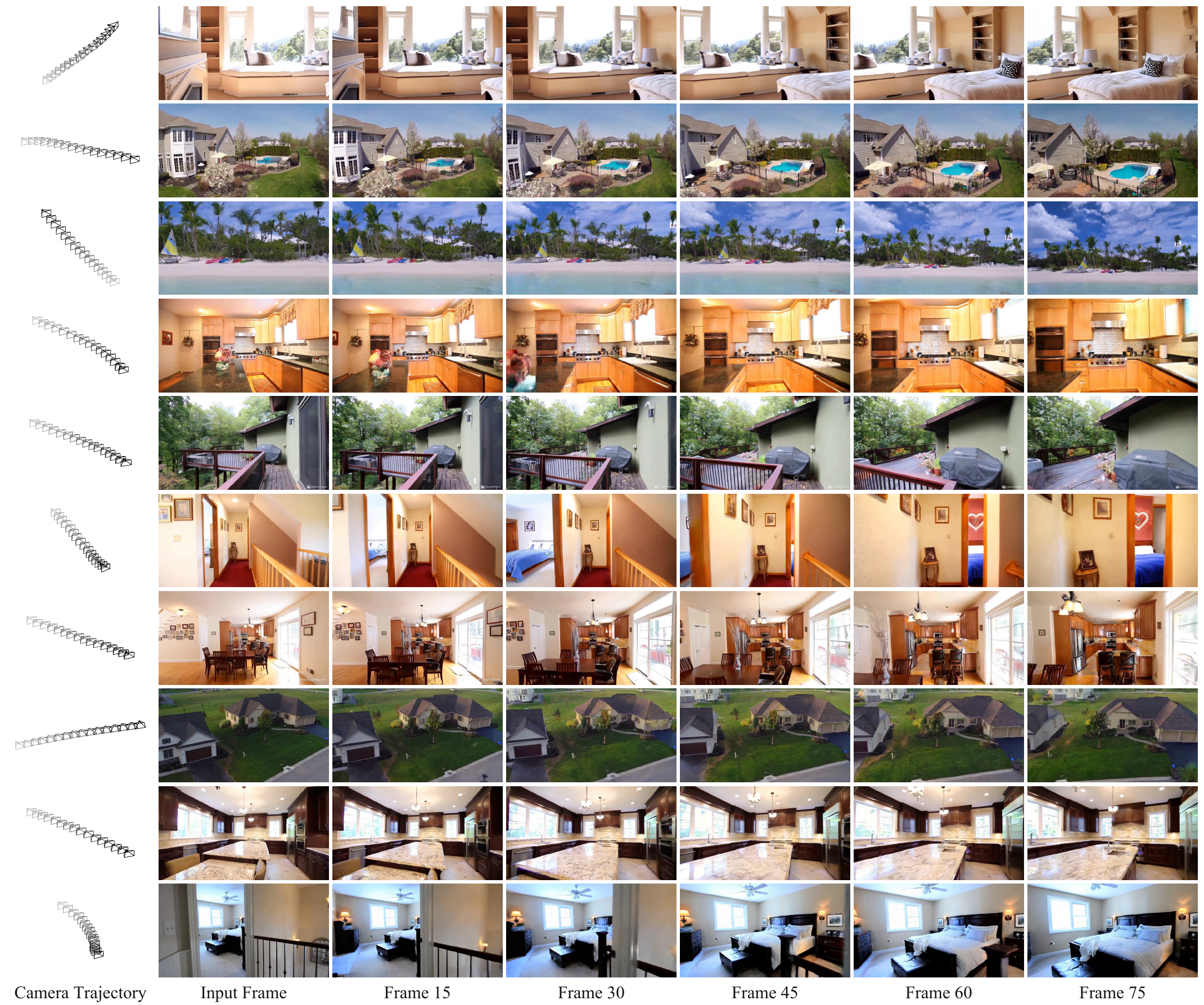}
\caption{\textbf{Long-horizon novel-view synthesis on RealEstate10K.}
Each row shows the camera trajectory, input image, and five generated views for one example. Camera markers progress from gray to black over time and use a fixed viewing orientation relative to each input camera.}
\label{fig:nvs-additional}
\end{figure}

\paragraph{Novel-view synthesis.}
Figure~\ref{fig:nvs-additional} presents ten examples of long-horizon generation on RealEstate10K. Given the input image and camera trajectory, \method\ generates subsequent views over three autoregressive chunks. We show frames 15, 30, 45, 60, and 75 to illustrate scene appearance and layout as the camera moves.

% \clearpage

\section{Additional Memory Details}

\paragraph{Memory Retention Across Successive Writes.}
We examine whether early scene content remains recoverable as new observations are written to HexMemory. On a 129-frame sequence, we successively write the initial frame and four ground-truth chunks, then repeatedly reconstruct chunk 1 from each resulting memory state. The input frame, camera trajectory, and prompt remain fixed across reconstructions. We exclude the fixed input frame when computing accuracy, allowing this probe to assess how subsequent writes affect the reconstruction of earlier content.

With memory disabled, chunk 1 reconstruction achieves 12.29\,dB PSNR. After chunk 1 enters memory at write 2, this improves to 16.71\,dB. Reconstruction PSNR remains 16.67, 16.35, and 16.30\,dB after writes 3, 4, and 5, respectively i.e., a total decline of just 0.41\,dB over three additional writes, averaging 0.14\,dB per write. Even after write 5, reconstruction remains 4.01\,dB above the memory-disabled baseline. Thus, in this sequence, earlier content remains recoverable with limited degradation while new observations are incorporated into the same fixed-size feature memory.

% \paragraph{Memory Storage.}
% Table~\ref{tab:memory} compares memory storage at
% matched history lengths. Spatia's RGB history and LSM-World's
% latent point cache grow as observations accumulate. \method's feature planes and confidence maps occupy 19.77\,MB. Their dimensions remain fixed as successive writes extend the represented spatial and temporal ranges.

% \begin{table}[!htbp]
% \caption{\textbf{Memory storage at matched history lengths.}
% Values are in decimal MB. We count RGB history for Spatia, latent features for LSM-World, and feature planes and confidence maps for \method.
% }
% \label{tab:memory}
% \centering
% \small
% \begin{tabular}{@{}lrrrrrr@{}}
% \toprule
% & \multicolumn{6}{c}{Storage (MB)$\downarrow$} \\
% \cmidrule(l){2-7}
% Stored RGB frames & 33 & 65 & 97 & 129 & 161 & 193 \\
% \midrule
% Spatia
% & 86.98 & 171.33 & 255.67 & 340.02 & 424.37 & 508.71 \\
% LSM-World
% & \textbf{7.22} & \textbf{12.99} & \textbf{18.76}
% & \underline{24.53} & \underline{30.31} & \underline{36.08} \\
% \method
% & \underline{19.77} & \underline{19.77} & \underline{19.77}
% & \textbf{19.77} & \textbf{19.77} & \textbf{19.77} \\
% \bottomrule
% \end{tabular}
% \end{table}

% \paragraph{Stored 3D Coordinates.}
% We additionally retain XYZ coordinates for geometric lookup, requiring 12 bytes per point in FP32. For a 65-frame evaluation, each run stores 59,840 points, occupying merely 0.72\,MB, or 3.6\% of the median HexMemory footprint. Under the same write schedule, the projected XYZ storage at 193 frames is just 2.07\,MB.

\paragraph{Explanation of Spatiotemporal Planes.}
Here, we provide further interpretation of HexMemory through visualizations of its spatiotemporal planes, spatial confidence maps, and stored features. The spatiotemporal planes $ZT$, $YT$, and $XT$ organize memory features along one spatial coordinate and the write-time coordinate $\tau$. In the visualizations, $\tau$ increases from bottom to top, and each horizontal slice summarizes the features stored at the corresponding write-time coordinate. The occupied extent indicates the range of observed scene content along that spatial axis.

In the alley sequence, the $ZT$ plane records the longitudinal extent of observations as the camera advances. Its lower-$z$ boundary shifts toward larger coordinates over time, consistent with nearby content leaving the forward-facing view. The $YT$ plane records the vertical extent of observations, displayed with height increasing from left to right. Its upper-height boundary generally contracts as the camera approaches the surrounding buildings and their upper portions leave the view. In this sequence, the maximum observed height remains approximately $0.37$ times the remaining observed distance along the alley, reflecting the camera's viewing geometry. This contraction eases near the end as the courtyard opens into view.

\begin{figure}[!htbp]
\centering
\includegraphics[width=\linewidth]{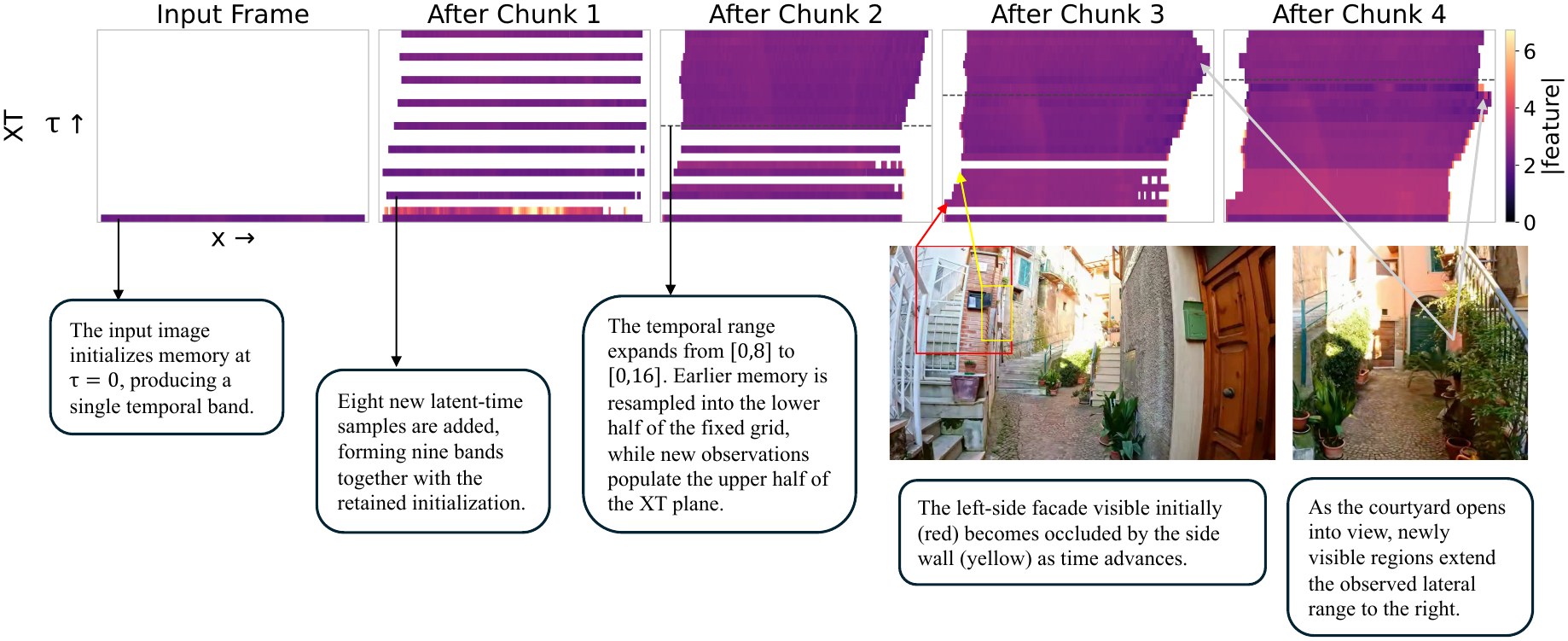}
\caption{\textbf{Temporal organization of the $XT$ feature plane.}
The input image initializes memory at $\tau = 0$, and the first chunk adds
eight new latent-time samples. Subsequent updates expand the temporal
range and resample the accumulated memory within a fixed-resolution
grid. Dashed lines mark the previous temporal extent. The annotated RGB views
illustrate changes in lateral visibility as the camera advances through
the alley.}
\label{fig:hexplane-timeplanes}
\end{figure}

The $XT$ plane records the lateral extent of observations
(Figure~\ref{fig:hexplane-timeplanes}), which remains approximately
$14$--$15$ world-coordinate units through much of the sequence,
consistent with the enclosing alley walls. Its right boundary extends
from approximately $x=4.5$ to $x=7.4$ as additional parts of the courtyard
become visible. Its left boundary moves inward by approximately one unit
as the nearby staircase and facade visible in the input image leave the
view. These changes illustrate how the spatiotemporal planes retain information about when different spatial regions were observed.

\begin{figure}[!htbp]
\centering
\includegraphics[width=\linewidth]{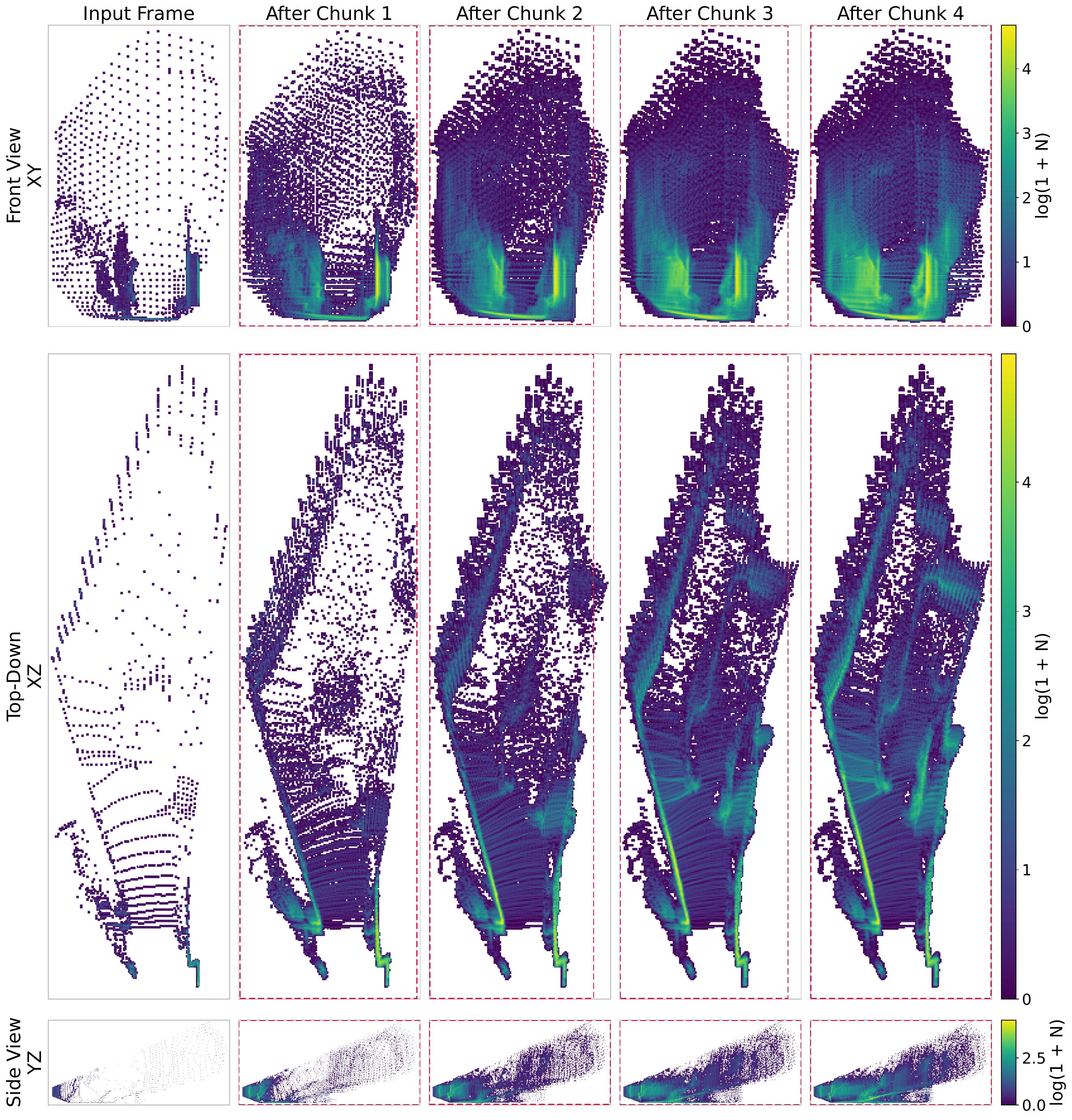}
\caption{\textbf{Progression of HexMemory confidence maps.}
Columns show initialization from the input frame and the memory state after each of four generated chunks. Colors represent $\log(1+N)$, where $N$ is the accumulated interpolation weight, with a shared color scale across updates for each plane.}
\label{fig:hexplane-evidence}
\end{figure}

\clearpage
\paragraph{Confidence-map visualization.}
Figure~\ref{fig:hexplane-evidence} visualizes the confidence maps associated with the spatial planes. These maps record the interpolation weights accumulated during memory writes and provide the weights used for fusion. We display them on a logarithmic scale to show where observations contribute to the memory and how this support evolves across updates.

\clearpage

\begin{figure}[!htbp]
\centering
\includegraphics[width=\linewidth]{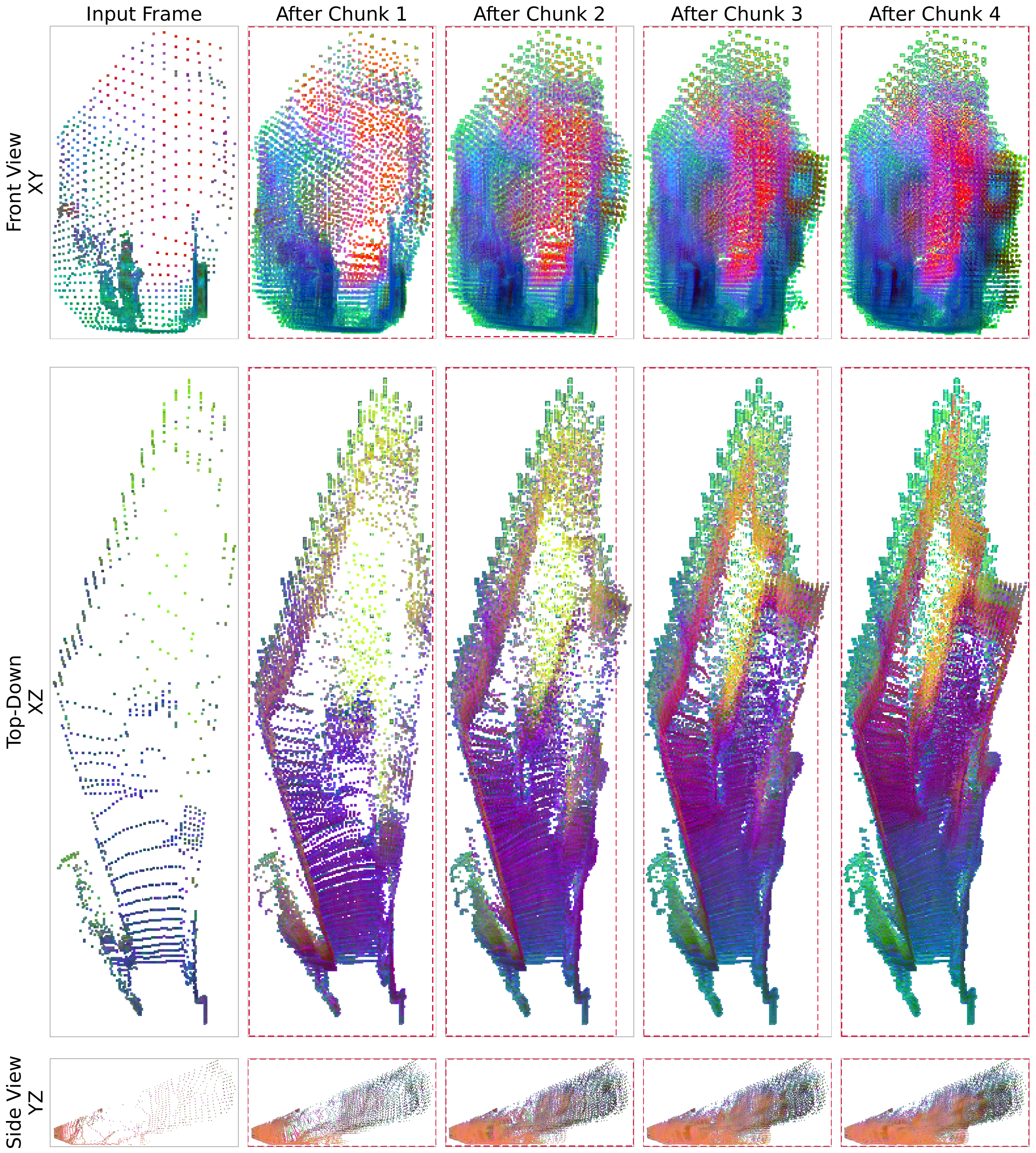}
\caption{\textbf{Spatial feature organization across memory updates.}
Columns show initialization from the input frame and the memory state after each of four generated chunks. The first three principal components of each plane's features are mapped to RGB, using a fixed PCA basis and color normalization derived from its final state. This allows feature patterns to be followed across updates within each plane.}
\label{fig:hexplane-pca}
\end{figure}

\paragraph{Feature visualization.}
Figure~\ref{fig:hexplane-pca} complements the feature-magnitude visualization in the main manuscript by displaying the spatial plane features through principal component analysis (PCA). For each plane, we compute a three-component PCA basis from the final memory state and apply the same projection across all displayed updates. Mapping these components to RGB provides a consistent view of feature organization as new observations are incorporated. The visualizations reveal spatially coherent feature patterns, including repeated banded structures and distinct regions that remain recognizable across successive updates as the observed scene coverage expands.

%% file: iclr2027_conference.bib
@String(PAMI  = {IEEE Trans. Pattern Anal. Mach. Intell.})

@String(CVPR  = {IEEE Conf. Comput. Vis. Pattern Recog.})

@String(ICCV  = {Int. Conf. Comput. Vis.})

@String(ECCV  = {Eur. Conf. Comput. Vis.})

@String(NeurIPS = {Adv. Neural Inform. Process. Syst.})

@String(ICML  = {Int. Conf. Mach. Learn.})

@String(ICLR  = {Int. Conf. Learn. Represent.})

@String(TMLR  = {Trans. Mach. Learn Res.})

@String(TOG   = {ACM Trans. Graph.})

@String(PAMI  = {IEEE TPAMI})

@String(CVPR  = {CVPR})

@String(ICCV  = {ICCV})

@String(ECCV  = {ECCV})

@String(TOG   = {ACM TOG})

@String(ICLR  = {ICLR})

@String(SIGGRAPH = {SIGGRAPH})

@String(ICML = {ICML})

@inproceedings{wang2024motionctrl,
  title = {{MotionCtrl}: A Unified and Flexible Motion Controller for Video Generation},
  author = {Wang, Zhouxia and Yuan, Ziyang and Wang, Xintao and Li, Yaowei and Chen, Tianshui and Xia, Menghan and Luo, Ping and Shan, Ying},
  booktitle = SIGGRAPH,
  year = {2024}
}

@inproceedings{he2025cameractrl,
  title = {{CameraCtrl}: Enabling Camera Control for Video Diffusion Models},
  author = {He, Hao and Xu, Yinghao and Guo, Yuwei and Wetzstein, Gordon and Dai, Bo and Li, Hongsheng and Yang, Ceyuan},
  booktitle = ICLR,
  year = {2025}
}

@article{xu2024camco,
  title = {{CamCo}: Camera-Controllable {3D}-Consistent Image-to-Video Generation},
  author = {Xu, Dejia and Nie, Weili and Liu, Chao and Liu, Sifei and Kautz, Jan and Wang, Zhangyang and Vahdat, Arash},
  journal = {arXiv preprint arXiv:2406.02509},
  year = {2024}
}

@article{zheng2024cami2v,
  title = {{CamI2V}: Camera-Controlled Image-to-Video Diffusion Model},
  author = {Zheng, Guangcong and Li, Teng and Jiang, Rui and Lu, Yehao and Wu, Tao and Li, Xi},
  journal = {arXiv preprint arXiv:2410.15957},
  year = {2024}
}

@article{chen2026memobench,
  title = {{MemoBench}: Benchmarking World Modeling in Dynamically Changing Environments},
  author = {Chen, Haoyu and Zhou, Kaichen and Hua, Hang and Zhang, Kaile and Qian, Jingwen and Ma, Wufei and Chen, Haonan and Liu, Chunjiang and Zhao, Yizhou and Wang, Xiaoyuan and others},
  journal = {arXiv preprint arXiv:2606.27537},
  year = {2026}
}

@inproceedings{bahmani2025vd3d,
  title = {{VD3D}: Taming Large Video Diffusion Transformers for {3D} Camera Control},
  author = {Bahmani, Sherwin and Skorokhodov, Ivan and Siarohin, Aliaksandr and Menapace, Willi and Qian, Guocheng and Vasilkovsky, Michael and Lee, Hsin-Ying and Wang, Chaoyang and Zou, Jiaxu and Tagliasacchi, Andrea and others},
  booktitle = ICLR,
  year = {2025}
}

@inproceedings{bahmani2025ac3d,
  title = {{AC3D}: Analyzing and Improving {3D} Camera Control in Video Diffusion Transformers},
  author = {Bahmani, Sherwin and Skorokhodov, Ivan and Qian, Guocheng and Siarohin, Aliaksandr and Menapace, Willi and Tagliasacchi, Andrea and Lindell, David B and Tulyakov, Sergey},
  booktitle = CVPR,
  year = {2025}
}

@inproceedings{zhang2026unified,
  title = {Unified Camera Positional Encoding for Controlled Video Generation},
  author = {Zhang, Cheng and Li, Boying and Wei, Meng and Cao, Yan-Pei and Gambardella, Camilo and Phung, Dinh and Cai, Jianfei},
  booktitle = CVPR,
  year = {2026}
}

@article{li2026rerope,
  title = {{ReRoPE}: Repurposing {RoPE} for Relative Camera Control},
  author = {Li, Chunyang and Yang, Yuanbo and Shao, Jiahao and Zhou, Hongyu and Schwarz, Katja and Liao, Yiyi},
  journal = {arXiv preprint arXiv:2602.08068},
  year = {2026}
}

@inproceedings{zhao2026cetcam,
  title={{CETCam}: Camera-Controllable Video Generation via Consistent and Extensible Tokenization},
  author={Zhao, Zelin and Gong, Xinyu and Liu, Bangya and Song, Ziyang and Zhang, Jun and Wu, Suhui and Chen, Yongxin and Zhang, Hao},
  booktitle={CVPR Findings},
  year = {2026}
}

@inproceedings{li2025realcam,
  title = {{RealCam-I2V}: Real-World Image-to-Video Generation with Interactive Complex Camera Control},
  author = {Li, Teng and Zheng, Guangcong and Jiang, Rui and Zhan, Shuigen and Wu, Tao and Lu, Yehao and Lin, Yining and Deng, Chuanyun and Xiong, Yepan and Chen, Min and others},
  booktitle = ICCV,
  year = {2025}
}

@inproceedings{ren2025gen3c,
  title = {{GEN3C}: {3D}-Informed World-Consistent Video Generation with Precise Camera Control},
  author = {Ren, Xuanchi and Shen, Tianchang and Huang, Jiahui and Ling, Huan and Lu, Yifan and Nimier-David, Merlin and M{\"u}ller, Thomas and Keller, Alexander and Fidler, Sanja and Gao, Jun},
  booktitle = CVPR,
  year = {2025}
}

@inproceedings{wang2026epic,
  title = {{EPiC}: Efficient Video Camera Control Learning with Precise Anchor-Video Guidance},
  author = {Wang, Zun and Cho, Jaemin and Li, Jialu and Lin, Han and Yoon, Jaehong and Zhang, Yue and Bansal, Mohit},
  booktitle = ICML,
  year = {2026}
}

@article{yu2024viewcrafter,
  title = {{ViewCrafter}: Taming Video Diffusion Models for High-Fidelity Novel View Synthesis},
  author = {Yu, Wangbo and Xing, Jinbo and Yuan, Li and Hu, Wenbo and Li, Xiaoyu and Huang, Zhipeng and Gao, Xiangjun and Wong, Tien-Tsin and Shan, Ying and Tian, Yonghong},
  journal = PAMI,
  year = {2025}
}

@inproceedings{he2025cameractrlii,
  title = {{CameraCtrl II}: Dynamic Scene Exploration via Camera-Controlled Video Diffusion Models},
  author = {He, Hao and Yang, Ceyuan and Lin, Shanchuan and Xu, Yinghao and Wei, Meng and Gui, Liangke and Zhao, Qi and Wetzstein, Gordon and Jiang, Lu and Li, Hongsheng},
  booktitle = ICCV,
  year = {2025}
}

@article{zhao2026geostream,
  title = {{GeoStream}: Toward Precise Camera Controlled Streaming Video Generation},
  author = {Zhao, Yizhou and Wang, Yifan and Wang, Xiaoyuan and Wu, Yushu and Zhang, Hao and Haji-Ali, Moayed and Abdal, Rameen and Mirzaei, Ashkan and Li, Yanyu and Menapace, Willi and others},
  journal = {arXiv preprint arXiv:2606.15162},
  year = {2026}
}

@inproceedings{zhou2023dynpoint,
  title = {{DynPoint}: Dynamic Neural Point for View Synthesis},
  author = {Zhou, Kaichen and Zhong, Jia-Xing and Shin, Sangyun and Lu, Kai and Yang, Yiyuan and Markham, Andrew and Trigoni, Niki},
  booktitle = NeurIPS,
  year = {2023}
}

@article{ye2026mind,
  title = {{MIND}: Benchmarking Memory Consistency and Action Control in World Models},
  author = {Ye, Yixuan and Lu, Xuanyu and Jiang, Yuxin and Gu, Yuchao and Zhao, Rui and Liang, Qiwei and Pan, Jiachun and Zhang, Fengda and Wu, Weijia and Wang, Alex Jinpeng},
  journal = {arXiv preprint arXiv:2602.08025},
  year = {2026}
}

@article{zhang2026mbench,
  title = {{MBench}: A Comprehensive Benchmark on Memory Capability for Video World Models},
  author = {Zhang, Shengjun and Zhang, Zhang and Huang, Simin and Tang, Zhenyu and Wang, Hanyang and Dai, Chensheng and Chen, Min and Li, Yifan and Li, Yuxin and Chen, Yingjie and others},
  journal = {arXiv preprint arXiv:2606.00793},
  year = {2026}
}

@article{ma2026out,
  title = {Out of Sight, Out of Mind? Evaluating State Evolution in Video World Models},
  author = {Ma, Ziqi and Liufu, Mengzhan and Gkioxari, Georgia},
  journal = {arXiv preprint arXiv:2603.13215},
  year = {2026}
}

@inproceedings{xiao2026worldmem,
  title = {{WorldMem}: Long-Term Consistent World Simulation with Memory},
  author = {Xiao, Zeqi and Lan, Yushi and Zhou, Yifan and Ouyang, Wenqi and Yang, Shuai and Zeng, Yanhong and Pan, Xingang},
  booktitle = NeurIPS,
  year = {2025}
}

@inproceedings{yu2025context,
  title = {Context as Memory: Scene-Consistent Interactive Long Video Generation with Memory Retrieval},
  author = {Yu, Jiwen and Bai, Jianhong and Qin, Yiran and Liu, Quande and Wang, Xintao and Wan, Pengfei and Zhang, Di and Liu, Xihui},
  booktitle = {Proceedings of the SIGGRAPH Asia 2025 Conference Papers},
  year = {2025}
}

@inproceedings{li2025vmem,
  title = {{VMem}: Consistent Interactive Video Scene Generation with Surfel-Indexed View Memory},
  author = {Li, Runjia and Torr, Philip and Vedaldi, Andrea and Jakab, Tomas},
  booktitle = ICCV,
  year = {2025}
}

@article{zhou2026stream3d,
  title = {{Stream3D}: Sequential Multi-View {3D} Generation via Evidential Memory},
  author = {Zhou, Kaichen and Bai, Zeyang and Chang, Xinhai and Wang, Mengyu and Liang, Paul and Zhan, Fangneng},
  journal = {arXiv preprint arXiv:2605.21472},
  year = {2026}
}

@inproceedings{xu2026ucm,
  title = {{UCM}: Unified Modeling of Camera Control and Memory with Time-Aware Positional Encoding Warping for World Models},
  author = {Xu, Tian-Xing and Wang, Zi-Xuan and Wang, Guangyuan and Hu, Li and Zhang, Zhongyi and Zhang, Peng and Zhang, Bang and Zhang, Song-Hai},
  booktitle = SIGGRAPH,
  year = {2026}
}

@article{yang2026decmem,
  title = {{DecMem}: Towards Minute-Long Consistent World Generation with Decoupled Memory},
  author = {Yang, Zhenhao and Wu, Xiaoshi and Lv, Zhengyao and Shi, Xiaoyu and Wang, Xintao and Wan, Pengfei and Gai, Kun and Wong, Kwan-Yee K},
  journal = {arXiv preprint arXiv:2605.31336},
  year = {2026}
}

@article{wei2026geometry,
  title = {Geometry-Aware Implicit Memory for Video World Models},
  author = {Wei, Zhengxuan and Guo, Xu and Li, Xinghui and Xiang, Xunzhi and Wei, Min and Zhu, Yiran and Wang, Qiulin and Wang, Xintao and Wan, Pengfei and Hou, Xiangwang and others},
  journal = {arXiv preprint arXiv:2606.02436},
  year = {2026}
}

@article{wu2026addressable,
  title = {Addressable Memory for Video World Models},
  author = {Wu, Xindi and Elflein, Sven and Lucas, James and Russakovsky, Olga and Leal-Taix{\'e}, Laura and Paschalidou, Despoina and Lorraine, Jonathan and O{\v{s}}ep, Aljo{\v{s}}a},
  journal = {arXiv preprint arXiv:2608.07408},
  year = {2026}
}

@inproceedings{wu2026video,
  title = {Video World Models with Long-Term Spatial Memory},
  author = {Wu, Tong and Yang, Shuai and Po, Ryan and Xu, Yinghao and Liu, Ziwei and Lin, Dahua and Wetzstein, Gordon},
  booktitle = NeurIPS,
  year = {2025}
}

@inproceedings{zhao2026spatia,
  title = {{Spatia}: Video Generation with Updatable Spatial Memory},
  author = {Zhao, Jinjing and Wei, Fangyun and Liu, Zhening and Zhang, Hongyang and Xu, Chang and Lu, Yan},
  booktitle = CVPR,
  year = {2026}
}

@article{yu2026mosaicmem,
  title = {{MosaicMem}: Hybrid Spatial Memory for Controllable Video World Models},
  author = {Yu, Wei and Qian, Runjia and Li, Yumeng and Wang, Liquan and Yin, Songheng and P, Sri Siddarth Chakaravarthy and Anthony, Dennis and Ye, Yang and Li, Yidi and Wan, Weiwei and others},
  journal = {arXiv preprint arXiv:2603.17117},
  year = {2026}
}

@article{wang2026latent,
  title = {Latent Spatial Memory for Video World Models},
  author = {Wang, Weijie and Zhao, Haoyu and Yang, Yifan and Chen, Feng and Zhang, Zeyu and He, Yefei and Duan, Zicheng and Chen, Donny Y and Yang, Yuqing and Zhuang, Bohan},
  journal = {arXiv preprint arXiv:2606.09828},
  year = {2026}
}

@inproceedings{cao2023hexplane,
  title = {{HexPlane}: A Fast Representation for Dynamic Scenes},
  author = {Cao, Ang and Johnson, Justin},
  booktitle = CVPR,
  year = {2023}
}

@inproceedings{fridovich2023k,
  title = {{K-Planes}: Explicit Radiance Fields in Space, Time, and Appearance},
  author = {Fridovich-Keil, Sara and Meanti, Giacomo and Warburg, Frederik Rahb{\ae}k and Recht, Benjamin and Kanazawa, Angjoo},
  booktitle = CVPR,
  year = {2023}
}

@article{lian2025loopnav,
  title={{LoopNav}: Benchmarking Spatial Consistency in World Models},
  author={Lian, Kewei and Cai, Shaofei and Liang, Yitao and Liu, Anji},
  journal={arXiv preprint arXiv:2505.22976},
  year={2025}
}

@inproceedings{song2025history,
  title={History-Guided video diffusion},
  author={Song, Kiwhan and Chen, Boyuan and Simchowitz, Max and Du, Yilun and Tedrake, Russ and Sitzmann, Vincent},
  booktitle=ICML,
  year={2025}
}

@article{oshima2025worldpack,
  title={WorldPack: Dynamic Frame Compression for Long-context Video World Modeling},
  author={Oshima, Yuta and Iwasawa, Yusuke and Suzuki, Masahiro and Matsuo, Yutaka and Furuta, Hiroki},
  journal=TMLR,
  year={2026}
}

@inproceedings{yu2024wonderjourney,
  title={{Wonderjourney}: Going from anywhere to everywhere},
  author={Yu, Hong-Xing and Duan, Haoyi and Hur, Junhwa and Sargent, Kyle and Rubinstein, Michael and Freeman, William T and Cole, Forrester and Sun, Deqing and Snavely, Noah and Wu, Jiajun and others},
  booktitle=CVPR,
  year={2024}
}

@inproceedings{yu2025wonderworld,
  title={{Wonderworld}: Interactive {3d} scene generation from a single image},
  author={Yu, Hong-Xing and Duan, Haoyi and Herrmann, Charles and Freeman, William T and Wu, Jiajun},
  booktitle=CVPR,
  year={2025}
}

@inproceedings{chen2024diffusion,
  title={Diffusion Forcing: Next-token Prediction Meets Full-Sequence Diffusion},
  author={Chen, Boyuan and Mart{\'\i} Mons{\'o}, Diego and Du, Yilun and Simchowitz, Max and Tedrake, Russ and Sitzmann, Vincent},
  booktitle=NeurIPS,
  year={2024}
}

@article{wang2025evoworld,
  title={{Evoworld}: Evolving panoramic world generation with explicit {3D} memory},
  author={Wang, Jiahao and Ye, Luoxin and Lu, TaiMing and Xiao, Junfei and Zhang, Jiahan and Guo, Yuxiang and Liu, Xijun and Chellappa, Rama and Peng, Cheng and Yuille, Alan and others},
  journal={arXiv preprint arXiv:2510.01183},
  year={2025}
}

@article{duan2026liveworld,
  title={{Liveworld}: Simulating out-of-sight dynamics in generative video world models},
  author={Duan, Zicheng and Xia, Jiatong and Zhang, Zeyu and Zhang, Wenbo and Zhou, Gengze and Gou, Chenhui and He, Yefei and Chen, Feng and Zhang, Xinyu and Liu, Lingqiao},
  journal={arXiv preprint arXiv:2603.07145},
  year={2026}
}

@article{lu2026current,
  title={Current World Models Lack a Persistent State Core},
  author={Lu, Jinpeng and Zhu, Dexu and Shi, Haoyuan and Cai, Linghan and Tang, Guo and Chen, Yinda and Cao, Jie and Tang, Duyu and Zhang, Yi and Dai, Yong and others},
  journal={arXiv preprint arXiv:2606.20545},
  year={2026}
}

@article{chen2026out,
  title={Out of sight but not out of mind: Hybrid memory for dynamic video world models},
  author={Chen, Kaijin and Liang, Dingkang and Zhou, Xin and Ding, Yikang and Liu, Xiaoqiang and Wan, Pengfei and Bai, Xiang},
  journal={arXiv preprint arXiv:2603.25716},
  year={2026}
}

@article{wan2025,
  title={Wan: Open and Advanced Large-Scale Video Generative Models},
  author={{Team Wan}},
  journal={arXiv preprint arXiv:2503.20314},
  year={2025}
}

@inproceedings{jiang2025vace,
  title={VACE: All-in-One Video Creation and Editing},
  author={Jiang, Zeyinzi and Han, Zhen and Mao, Chaojie and Zhang, Jingfeng and Pan, Yulin and Liu, Yu},
  booktitle=ICCV,
  year={2025}
}

@inproceedings{hu2022lora,
  title={LoRA: Low-Rank Adaptation of Large Language Models},
  author={Hu, Edward J. and Shen, Yelong and Wallis, Phillip and Allen-Zhu, Zeyuan and Li, Yuanzhi and Wang, Shean and Wang, Lu and Chen, Weizhu},
  booktitle=ICLR,
  year={2022}
}

@article{huang2025vipe,
  title={ViPE: Video Pose Engine for 3D Geometric Perception},
  author={Huang, Jiahui and Zhou, Qunjie and Rabeti, Hesam and Korovko, Aleksandr and Ling, Huan and Ren, Xuanchi and Shen, Tianchang and Gao, Jun and others},
  journal={arXiv preprint arXiv:2508.10934},
  year={2025}
}

@inproceedings{duan2025worldscore,
  title={WorldScore: A Unified Evaluation Benchmark for World Generation},
  author={Duan, Haoyi and Yu, Hong-Xing and Chen, Sirui and Fei-Fei, Li and Wu, Jiajun},
  booktitle=ICCV,
  year={2025}
}

@article{zhou2018stereo,
  title = {Stereo Magnification: Learning View Synthesis using Multiplane Images},
  author = {Zhou, Tinghui and Tucker, Richard and Flynn, John and Fyffe, Graham and Snavely, Noah},
  journal = TOG,
  year = {2018}
}

@inproceedings{lipman2023flow,
  title={Flow Matching for Generative Modeling},
  author={Lipman, Yaron and Chen, Ricky T. Q. and Ben-Hamu, Heli and Nickel, Maximilian and Le, Matt},
  booktitle=ICLR,
  year={2023}
}

@inproceedings{yu2026memlearner,
  title={MemLearner: Learning to Query Context Memory for Video World Models},
  author={Yu, Jiwen and Gao, Jianxiong and Bai, Jianhong and Qin, Yiran and Huang, Kaiyi and Liu, Quande and Wang, Xintao and Wan, Pengfei and Gai, Kun and Liu, Xihui},
  booktitle=ECCV,
  year={2026}
}

@article{kim2026memrope,
  title={Memrope: Training-free infinite video generation via evolving memory tokens},
  author={Kim, Youngrae and Hu, Qixin and Kuo, C-C Jay and Beerel, Peter A},
  journal={arXiv preprint arXiv:2603.12513},
  year={2026}
}

@article{xue2026ring,
  title={Ring Forcing: Towards Precise Long-Term Memory for Autoregressive Video Diffusion},
  author={Xue, Bowen and Feng, Brandon Y and Lin, Chenguo and Lin, Yuchen and Zeng, Yujia and Zhang, Lvmin and Agrawala, Maneesh and Yan, Honglei and Pan, Panwang},
  journal={arXiv preprint arXiv:2608.26794},
  year={2026}
}

@article{wang2026anchorweave,
  title={Anchorweave: World-consistent video generation with retrieved local spatial memories},
  author={Wang, Zun and Lin, Han and Yoon, Jaehong and Cho, Jaemin and Zhang, Yue and Bansal, Mohit},
  journal={arXiv preprint arXiv:2602.14941},
  year={2026}
}

@inproceedings{lee20263d,
  title={3d scene prompting for scene-consistent camera-controllable video generation},
  author={Lee, JoungBin and Jung, Jaewoo and Han, Jisang and Narihira, Takuya and Fukuda, Kazumi and Seo, Junyoung and Hong, Sunghwan and Mitsufuji, Yuki and Kim, Seungryong},
  booktitle=ICLR,
  year={2026}
}

@article{xu2026wonder,
  title={Wonder: Video World Model Done Better},
  author={Xu, Jiacong and Jiang, Hanwen and Shu, Zhixin and Sunkavalli, Kalyan and Patel, Vishal M and Mei, Yiqun},
  journal={arXiv preprint arXiv:2607.26037},
  year={2026}
}

@article{chen2026reworld,
  title={ReWorld: An Interactive World Model with Long-Horizon Memory},
  author={Chen, Zhifei and Wang, Luozhou and Shen, Guibao and Yan, Dongyu and Yang, Shuai and Xu, Tianshuo and Du, Yihua and Wang, Wei and Gui, Tianyi and Huang, Lianghua and others},
  journal={arXiv preprint arXiv:2608.23565},
  year={2026}
}

@article{wu2026infinite,
  title={Infinite-world: Scaling interactive world models to 1000-frame horizons via pose-free hierarchical memory},
  author={Wu, Ruiqi and He, Xuanhua and Cheng, Meng and Yang, Tianyu and Zhang, Yong and Kang, Zhuoliang and Cai, Xunliang and Wei, Xiaoming and Guo, Chunle and Li, Chongyi and others},
  journal={arXiv preprint arXiv:2602.02393},
  year={2026}
}

@article{yi2026worldkv,
  title={WorldKV: Efficient World Memory with World Retrieval and Compression},
  author={Yi, Jung and Kim, Minjae and Cho, Paul Hyunbin and Jang, Wooseok and Yun, Sangdoo and Kim, Seungryong},
  journal={arXiv preprint arXiv:2605.22718},
  year={2026}
}

@inproceedings{garcin2026persist,
  title={Beyond pixel histories: World models with persistent 3d state},
  author={Garcin, Samuel and Walker, Thomas and McDonagh, Steven and Pearce, Tim and Bilen, Hakan and He, Tianyu and Wang, Kaixin and Bian, Jiang},
  booktitle=ICML,
  year = {2026}
}

@article{guo2026memorize,
  title={Memorize When Needed: Decoupled Memory Control for Spatially Consistent Long-Horizon Video Generation},
  author={Guo, Yanjun and Zhang, Zhengqiang and Wang, Pengfei and Liang, Xinyue and Ma, Zhiyuan and Zhang, Lei},
  journal={arXiv preprint arXiv:2604.18215},
  year={2026}
}

@inproceedings{teed2020raft,
  title={Raft: Recurrent all-pairs field transforms for optical flow},
  author={Teed, Zachary and Deng, Jia},
  booktitle = ECCV,
  year={2020}
}

@inproceedings{zhang2025frame,
  title={Frame Context Packing and Drift Prevention in Next-Frame-Prediction Video Diffusion Models},
  author={Zhang, Lvmin and Cai, Shengqu and Li, Muyang and Wetzstein, Gordon and Agrawala, Maneesh},
  booktitle=NeurIPS,
  year={2025}
}

@article{huang2025voyager,
  title = {{Voyager}: Long-Range and World-Consistent Video Diffusion for Explorable {3D} Scene Generation},
  author = {Huang, Tianyu and Zheng, Wangguandong and Wang, Tengfei and Liu, Yuhao and Wang, Zhenwei and Wu, Junta and Jiang, Jie and Li, Hui and Lau, Rynson W. H. and Zuo, Wangmeng and Guo, Chunchao},
  journal = TOG,
  year = {2025}
}

@article{yu2025videossm,
  title={Videossm: Autoregressive long video generation with hybrid state-space memory},
  author={Yu, Yifei and Wu, Xiaoshan and Hu, Xinting and Hu, Tao and Sun, Yangtian and Lyu, Xiaoyang and Wang, Bo and Ma, Lin and Ma, Yuewen and Wang, Zhongrui and others},
  journal={arXiv preprint arXiv:2512.04519},
  year={2025}
}

@inproceedings{liu2026driveva,
  title={Driveva: Video action models are zero-shot drivers},
  author={Liu, Mengmeng and Zhang, Diankun and Liu, Jiuming and Cui, Jianfeng and Xie, Hongwei and Chen, Guang and Ye, Hangjun and Yang, Michael Ying and Nex, Francesco and Cheng, Hao},
  booktitle=ECCV,
  year={2026}
}

@article{liu2026universe,
  title={UNIVERSE: Unified Video Action Models for Autonomous Driving with Flexible Mask-Modulated Modality Generation},
  author={Liu, Mengmeng and Zhang, Diankun and Liu, Jiuming and Cui, Jianfeng and Xie, Hongwei and Chen, Guang and Ye, Hangjun and Nex, Francesco and Cheng, Hao and Yang, Michael Ying},
  journal={arXiv preprint arXiv:2607.05133},
  year={2026}
}

@article{zhang2026world,
  title={World Action Planner: Generalizable Decision-Making with Action-Conditioned World Models},
  author={Zhang, Xiangcheng and Du, Yilun},
  journal={arXiv preprint arXiv:2607.27599},
  year={2026}
}
